\documentclass[journal,twoside]{IEEEtran}

\usepackage{amsfonts}
\usepackage{amsmath}
\usepackage{xspace}
\usepackage{graphicx}
\usepackage{mathabx}
\usepackage{tabulary}
\usepackage{multirow}
\usepackage{makecell}
\usepackage{booktabs}
\usepackage{threeparttable}
\usepackage{enumitem}
\usepackage{url}
\usepackage{cite}
\usepackage{array}
\usepackage{color}
\usepackage{xcolor}
\usepackage{hyperref}
\usepackage{stfloats}
\usepackage{flushend} 

\newcommand{\system}{{MGMN}\xspace}
\newcommand{\bfsystem}{\textbf{{MGMN}}\xspace}
\newcommand{\tabincell}[2]{\begin{tabular}{@{}#1@{}}#2\end{tabular}}
\newcommand{\rpm}{\raisebox{.4ex}{$ \scriptstyle\pm $}}
\newcommand{\ie}{\textit{i}.\textit{e}.}
\newcommand{\eg}{\textit{e}.\textit{g}.}

\begin{document}
%
\title{Multilevel Graph Matching Networks for Deep Graph Similarity Learning}
%
%
%


\author{Xiang~Ling\IEEEauthorrefmark{1},
        Lingfei~Wu\IEEEauthorrefmark{1},~\IEEEmembership{Member,~IEEE,}
        Saizhuo~Wang,
        Tengfei~Ma,
        Fangli~Xu,
        Alex~X.~Liu,~\IEEEmembership{Fellow,~IEEE,}
        Chunming~Wu\IEEEauthorrefmark{2},
        and~Shouling~Ji\IEEEauthorrefmark{2},~\IEEEmembership{Member,~IEEE}
\thanks{This work was partly supported by
the National Key R\&D Program of China under No. 2020YFB1804705,
NSFC under No. 61772466 and U1936215,
the Key R\&D Program of Zhejiang Province under No. 2021C01036 and 2020C01021,
the Zhejiang Provincial Natural Science Foundation for Distinguished Young Scholars under No. LR19F020003,
the Major Scientific Project of Zhejiang Lab under No. 2018FD0ZX01,
and the Fundamental Research Funds for the Central Universities (Zhejiang University NGICS Platform).}%
\thanks{Xiang~Ling, Saizhuo~Wang, and Chunming~Wu are with the College of Computer Science and Technology, Zhejiang University, Hangzhou 310027, China (e-mail: \{lingxiang, szwang, wuchunming\}@zju.edu.cn);
Lingfei~Wu is with the JD.COM Silicon Valley Research Center, CA 94043, USA (e-mail: lwu@email.wm.edu);
Tengfei~Ma is with the IBM T. J. Watson Research Center, NY 10598, USA (e-mail: Tengfei.Ma1@ibm.com); 
Fangli~Xu is with the Squirrel AI Learning, NY, USA (e-mail: lili@yixue.us);
Alex~X.~Liu is with the Ant Financial Services Group, Hangzhou 310013, China (e-mail: alexliu@antfin.com);
Shouling~Ji is with the College of Computer Science and Technology, Zhejiang University, and the Binjiang Institute of Zhejiang University, Hangzhou, Zhejiang 310027, China (e-mail: sji@zju.edu.cn)}%
\thanks{Xiang~Ling\IEEEauthorrefmark{1} and Lingfei~Wu\IEEEauthorrefmark{1} contribute equally to this research.}%
\thanks{Chunming Wu\IEEEauthorrefmark{2} and Shouling Ji\IEEEauthorrefmark{2} are the co-corresponding authors.}%
}

%
%

\markboth{IEEE Transactions on Neural Networks and Learning Systems, 2021}%
{Ling \MakeLowercase{\textit{et al.}}: Multilevel Graph Matching Networks for Deep Graph Similarity Learning}
%



\maketitle

\begin{abstract}
While the celebrated graph neural networks yield effective representations for individual nodes of a graph, there has been relatively less success in extending to the task of graph similarity learning.
Recent work on graph similarity learning has considered either global-level graph-graph interactions or low-level node-node interactions, however ignoring the rich cross-level interactions (\eg, between each node of one graph and the other whole graph).
In this paper, we propose a multilevel graph matching network (\system) framework for computing the graph similarity between any pair of graph-structured objects in an end-to-end fashion.
In particular, the proposed \system consists of a node-graph matching network for effectively learning \emph{cross-level interactions} between each node of one graph and the other whole graph, and a siamese graph neural network to learn \emph{global-level interactions} between two input graphs.
Furthermore, to compensate for the lack of standard benchmark datasets, we have created and collected a set of datasets for both the graph-graph classification and graph-graph regression tasks with different sizes in order to evaluate the effectiveness and robustness of our models. 
Comprehensive experiments demonstrate that \system consistently outperforms state-of-the-art baseline models on both the graph-graph classification and graph-graph regression tasks.
Compared with previous work, \system also exhibits stronger robustness as the sizes of the two input graphs increase.
\end{abstract}

%
\IEEEpeerreviewmaketitle

\section{Introduction}\label{sec:intro}
\IEEEPARstart{L}{earning} 
a general similarity metric between arbitrary pairs of graph-structured objects is one of the key challenges in machine learning.
Such a learning problem often arises in a variety of real-world applications, ranging from graph similarity searching in graph-based databases~\cite{yan2002gspan}, to fewshot 3D action recognition~\cite{guo2018neural}, unknown malware detection~\cite{wangheterogeneous} and natural language processing \cite{chen2019reinforcement}, to name just a few.
Conceptually, classical \textit{exact} and \textit{error-tolerant} (\ie, inexact) graph matching techniques~\cite{bunke1983inexact,caetano2009learning,riesen2010exact,dwivedi2018error} provide a strong tool for learning a graph similarity metric.
For \textit{exact} graph matching, a strict one-to-one correspondence is required between nodes and edges of two input graphs, whereas \textit{error-tolerant} graph matching techniques attempt to compute a similarity score between two input graphs\cite{GNNBook-ch13-ling}.
However, in some real-world applications, the constraints of \textit{exact} graph matching techniques are too rigid (\eg, presence of noises or distortions in graphs, neglect of node features, no need for strict one-to-one correspondences, etc.).
Thus, in this paper, we focus on the \textit{error-tolerant} graph matching -- the graph similarity problem that learns a similarity score between a pair of input graphs.
Specifically, we consider the graph similarity problem as to learn a mapping between a pair of input graphs $(G^1, G^2) \in \mathcal{G} \times \mathcal{G}$ and a similarity score $y \in \mathcal{Y}$, based on a set of training triplets $(G^1_1, G^2_1, y_1), \cdots ,(G^1_n, G^2_n, y_n) \in \mathcal{G} \times \mathcal{G} \times \mathcal{Y}$ drawn from some fixed but unknown probability distribution in real-world applications.

Recent years have seen a surge of interests in graph neural networks (GNNs), which have been demonstrated to be a powerful class of deep learning models for learning node representations of graphs~\cite{bronstein2017geometric, wu2020comprehensive}.
Various GNN models have since been developed for learning effective node embedding vectors for the node classification task~\cite{kipf2016semi,GraphSage:hamilton2017inductive,velivckovic2017graph,chen2020iterative},
pooling the learned node embedding vectors into a graph-level embedding vector for the general graph classification task~\cite{ying2018hierarchical,ma2019graph,lee2019self,gao2019graph},
or combining with variational auto-encoder to learn the graph distribution for the graph generation task~\cite{simonovsky2018graphvae, li2018learning, samanta2018designing, you2018graphrnn}.
However, there is less study on learning the similarity score between two graphs using GNNs.
To learn a similarity score between two input graphs, a simple yet straightforward way is to first encode each graph as a vector of graph-level embedding via GNNs and then combine the two vectors of both input graphs to make a decision.
This approach is useful since it explores graph-level graph-graph interaction features that contain important information of the two input graphs for the graph similarity learning.
One obvious limitation of this approach lies in the fact of the ignorance of more fine-grained interaction features among different levels of embeddings between two input graphs.

Very recently, a few attempts of graph matching networks have been made to take into account low-level node-node interactions either by considering the histogram information or spatial patterns (using convolutional neural network~\cite{krizhevsky2012imagenet}) of the node-wise similarity matrix of node embeddings~\cite{bai2019simgnn, bai2020learning},
or by improving the node embeddings of one graph via incorporating the implicit attentive neighbors of the other graph~\cite{li2019graph}.
However, there are two significant challenges making these graph matching networks potentially ineffective:
i) how to effectively learn richer cross-level interactions between pairs of input graphs;
ii) how to integrate a multilevel granularity (\ie, both cross-level and global-level) of interactions between pairs of input graphs for computing the graph similarity in an end-to-end fashion.

In order to address the aforementioned challenges, in this paper, we propose a \textbf{M}ulti-Level \textbf{G}raph \textbf{M}atching \textbf{N}etwork (\system) for computing the graph similarity between any pairs of graph-structured objects in an end-to-end fashion.
In particular, \system consists of a novel node-graph matching network (NGMN) for effectively learning \emph{cross-level interaction features} by comparing each contextual node embedding of one graph against the attentive graph-level embedding of another graph,
and a siamese graph neural network (SGNN) for learning \emph{global-level interaction features} between two graphs.
Our final small prediction network leverages the multilevel granularity features that are learned from both cross-level and global-level interactions to perform either the graph-graph classification task or the graph-graph regression task.

The recently proposed graph matching networks~\cite{li2019graph, bai2019simgnn, bai2020learning} only compute graph similarity scores by considering either the graph-graph classification task (with a binary similarity label $y \in \{-1, 1\}$)~\cite{li2019graph}, or the graph-graph regression task (with a similarity score $y \in (0, 1]$)~\cite{bai2019simgnn,bai2020learning}.
It is noted that, the graph-graph classification task here is basically different from the general graph classification task~\cite{ying2018hierarchical,ma2019graph} that only assigns each individual graph with a label.
In contrast, the graph-graph classification task in our paper learns a binary similarity label (\ie, similar or dissimilar) for \textit{two input graphs} instead of \textit{one input graph}.
To demonstrate the effectiveness of our full model \system, we systematically investigate the performance of \system compared with these recently proposed graph matching networks on four benchmark datasets for both the graph-graph classification and graph-graph regression tasks.

Another important aspect is previous work does not consider the impact of the size of pairs of input graphs, which often plays an important role in determining the robustness performance of graph matching networks.
Motivated by this observation, we consider three different ranges of graph sizes (\ie, [3, 200], [20, 200], and [50, 200]) in order to evaluate the robustness of each graph matching model.
In addition, to compensate for the lack of standard benchmark datasets for the task of graph similarity learning, we create one new dataset from a real-world application together with a previously released dataset by~\cite{xu2017neural} for the graph-graph classification task.
Our code and data are available for research purposes at \url{https://github.com/kleincup/MGMN}.
In brief, we highlight our main contributions as follows:
\begin{itemize}
\item We first propose a novel node-graph matching network (NGMN) for effectively capturing the rich cross-level interaction features by comparing each contextual node embedding of one graph against the attentive graph-level embedding of another graph with the defined multi-perspective matching function.

\item We further present a multilevel graph matching network (\system) framework to compute the graph similarity between any pairs of graphs in an end-to-end fashion.
In particular, \system takes into account both cross-level and graph-level interactions between two input graphs.

\item We systematically investigate different factors on the performance of all graph matching networks such as different graph-graph similarity tasks (the graph-graph classification and graph-graph regression tasks) and different sizes of input graphs.

\item Comprehensive experiments demonstrate that \system consistently outperforms state-of-the-art baselines for both the graph-graph classification and graph-graph regression tasks.
Compared with previous work, the proposed \system also exhibits stronger robustness as the size of the two input graph increase.
\end{itemize}

\textbf{Roadmap}.
The remainder of this paper is organized as follows.
We briefly introduce the formulation of the graph similarity learning problem in Section~\ref{sec:problem} and describe the proposed \system model for computing the graph similarity between two graphs in Section~\ref{sec:model_overview}.
The performance of the \system model is systematically evaluated and analyzed in Section~\ref{sec:experiment}.
Section~\ref{sec:related_work} surveys related work and Section~\ref{sec:conclusion_future_work} finally concludes this work.
\section{Problem Formulation}\label{sec:problem}
In this section, we briefly introduce the problem formulation as follows.
Given a pair of input graphs $(G^1, G^2)$, the aim of graph similarity learning in this paper is to produce a similarity score $y = s(G^1, G^2) \in \mathcal{Y}$.
In particular, the graph $G^1 = (\mathcal{V}^1, \mathcal{E}^1)$ is represented as a set of $N$ nodes $v_i \in \mathcal{V}^1$ with a feature matrix $X^1 \in \mathcal{R}^{N \times d}$, edges $(v_i, v_{i'}) \in \mathcal{E}^1$ (binary or weighted) formulating an adjacency matrix $A^1 \in \mathcal{R}^{N \times N}$, and a degree matrix $\widetilde{D}^{1}_{ii} = \sum_{i'} A^1_{ii'}$.
Similarly,
the graph $G^2 = (\mathcal{V}^2, \mathcal{E}^2)$ is represented as a set of $M$ nodes $v_j \in \mathcal{V}^2$ with a feature matrix $X^2 \in \mathcal{R}^{M \times d}$, edges $(v_j, v_{j'}) \in \mathcal{E}^2$ (binary or weighted) formulating an adjacency matrix $A^2 \in \mathcal{R}^{M \times M}$, and a degree matrix $\widetilde{D}^{2}_{jj'} = \sum_{j'} A^2_{jj'}$.
It is noted that, when performing the graph-graph classification task, $y$ is a binary similarity label $y \in \mathcal{Y} = \{-1, 1\}$;
when performing the graph-graph regression task, $y$ is the continuous similarity score $y \in \mathcal{Y} = (0, 1]$.
We train our models based on a set of training triplet of input graph pairs and a scalar output score $(G^1_1,G^2_1,y_1), \cdots ,(G^1_n,G^2_n,y_n) \in \mathcal{G} \times \mathcal{G} \times \mathcal{Y}$ drawn from some fixed but unknown probability distribution in real-world applications.
A summary of important symbols and notations used in this paper can be found in Table~\ref{tab:notation}.

\begin{table}[t]
\centering
\caption{Important symbols and notations}
\label{tab:notation}
\renewcommand\tabcolsep{2.5pt}
\begin{tabular}{cl}
  \toprule
    \textbf{Symbols}&   \textbf{Definitions or Descriptions}\\
    \midrule
    $G^l, \ l=\{1, 2\}$                         & The $l$-th input graph, \ie, $G^1$ and $G^2$.         \\
    $N$                                         & The number of nodes in $G^1$ or the size of $G^1$.    \\
    $M$                                         & The number of nodes in $G^2$ or the size of $G^2$.    \\
    $\vec{h}^l_i, \ l=\{1, 2\}$                 & The hidden embedding of the $i$-th node in $G^l$.     \\
    $H^l, \ l=\{1, 2\}$                         & \tabincell{l}{The set of all node embeddings in $G^l$,\\\ie, $H^l = \{\vec{h}^l_i\}_{i=1}^{\{N,M\}}, \ l=\{1, 2\}$.} \\
    $\widetilde{\vec{h}}_G^l, \ l=\{1, 2\}$     & The graph-level embedding vector of $G^l$ from NGMN.  \\
    $\vec{h}_G^l, \ l=\{1, 2\}$                 & The graph-level embedding vector of $G^l$ from SGNN.  \\
  \bottomrule
\end{tabular}
\end{table}
\section{Multilevel Graph Matching Network}\label{sec:model_overview}
\begin{figure*}[ht]
    \centering
    \includegraphics[keepaspectratio=true, width=0.8\linewidth]{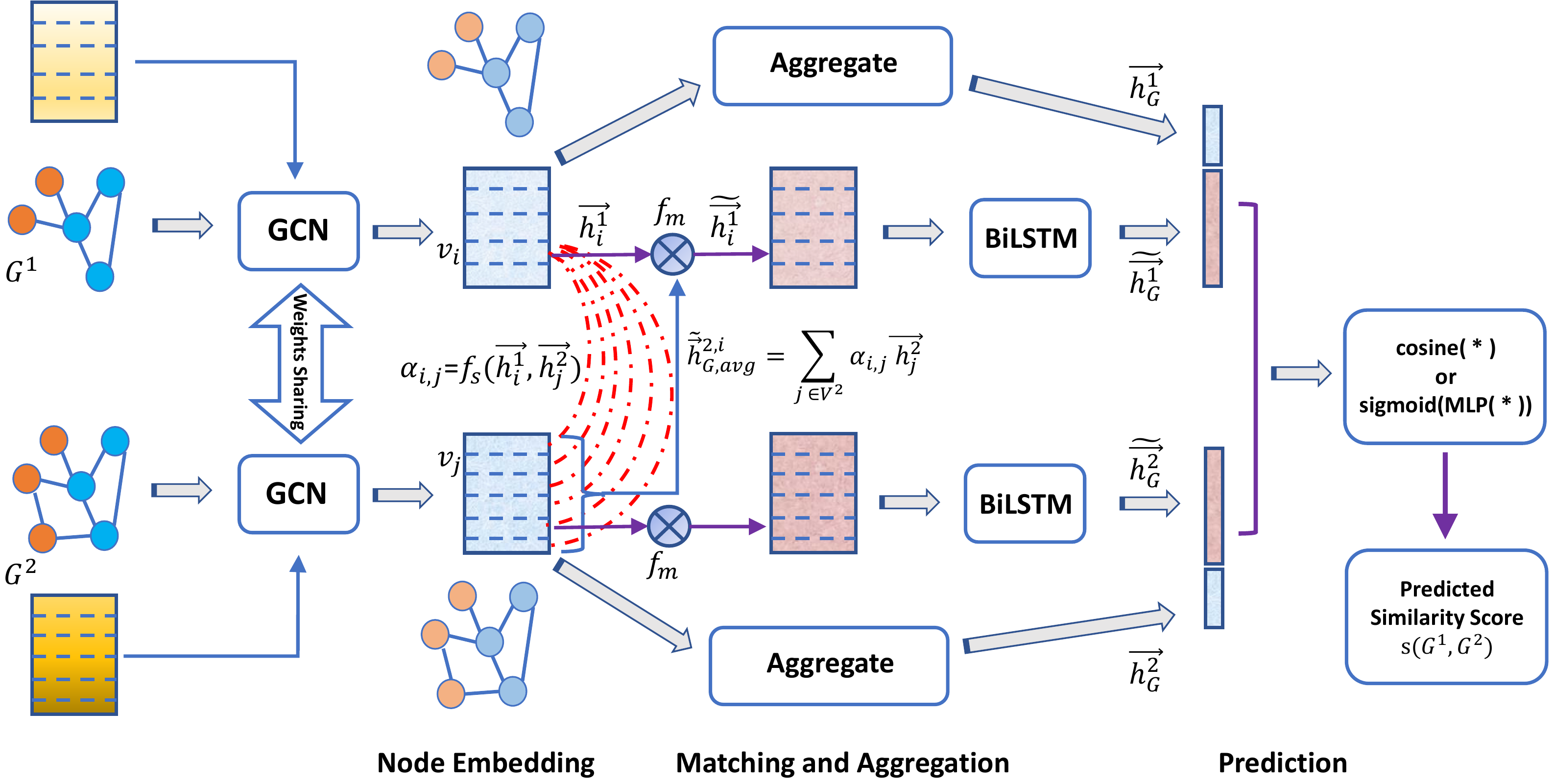}
    \caption{Overview architecture of the full model \system, consisting of two partial models: SGNN and NGMN. The two input graphs first go through NGMN and SGNN, resulting in aggregated graph-level embedding vectors after their corresponding aggregation layers. After that, we concatenate the two aggregated graph-level embedding vectors for each graph $G^l$, where one vector $\widetilde{\vec{h}}_G^l$ (long \& pink) is from NGMN and another vector $\vec{h}_G^l$ (short \& blue) is from SGNN, and then the two concatenated embedding vectors into the following prediction layer.}
    \label{fig:hgmn_model}
\end{figure*}

In this section, we detail the proposed multilevel graph matching network (\system) framework, which consists of a node-graph matching network (NGMN) and a siamese graph neural network (SGNN).
The overall model architecture of \system is shown in Fig.~\ref{fig:hgmn_model}.
In the following subsections,
we first introduce NGMN for effectively learning the cross-level node-graph interaction features between each node of one graph and the other whole graph, and then outline SGNN for learning the global-level interaction features between the two graphs.
Finally, we present our full model \system that combines NGMN and SGNN to learn both cross-level node-graph interactions as well as graph-level graph-graph interactions.

\subsection{NGMN for Cross-Level Interaction Learning}\label{subsec:ngmn}
Existing work has considered either global-level graph-graph interactions or low-level node-node interactions, ignoring the rich cross-level interactions between two input graphs.
Inspired by these observations, we propose a novel node-graph matching network (NGMN) to effectively learn the cross-level node-graph interaction features by comparing each contextual node embedding of one graph against the attentive graph-level embedding of another graph.
In general, NGMN consists of four layers:
1) node embedding layer;
2) node-graph matching layer;
3) aggregation layer;
and 4) prediction layer.
We will illustrate each layer in detail as follows.

\subsubsection{Node Embedding Layer}\label{subsubsec:ngmn_node_embedding}
The siamese network architecture~\cite{bromley1994signature} has achieved great success in many metric learning tasks such as visual recognition~\cite{bertinetto2016fully,varior2016gated}, video segmentation~\cite{shin2017pixel,lu2019see} and sentence similarity analysis~\cite{he2015multi,mueller2016siamese}.
In this layer, we consider a multi-layer graph convolution network (GCN)~\cite{kipf2016semi} with the siamese network architecture to generate node embeddings $H^l = \{\vec{h}^l_i\}_{i=1}^{\{N,M\}} \in \mathcal{R}^{\{N,M\} \times d'}$ for all nodes in either $G^1$ or $G^2$ as follows.
\begin{align}\label{eq:gcn}
\resizebox{.91\linewidth}{!}{$
  \displaystyle
    H^l = \sigma \Big(\widebar{A}^l \dots \sigma \Big( \widebar{A}^l \ \sigma \Big(\widebar{A}^l X^l W^{(0)}\Big) W^{(1)}\Big) \dots  W^{(T-1)}\Big), \; l = \{1, 2\}
$}
\end{align}
Here,
$\sigma$ is the activation function;
$\widebar{A}^l = (\widetilde{D}^l)^{-\frac{1}{2}} \widetilde{A}^l (\widetilde{D}^l)^{-\frac{1}{2}}$ is the normalized Laplacian matrix for $\widetilde{A}^l = A^l + I_{\{N,M\}}$ depending on $G^1$ or $G^2$;
$N$ and $M$ denote the number of nodes for both $G^1$ and $G^2$;
$H^l$ is the set of all learned node embeddings of the graph $G^l$;
$l=\{1, 2\}$ in the superscript of $H^l$, $\widebar{A}^l$, $X^l$ indicates it belongs to $G^1$ or $G^2$;
$T$ is the number of GCN layers;
$W^{(t)}, t \in \{0, 1, \dots, T-1\}$ is the layer-specific trainable weighted matrix of the $t$-th GCN layer.
It is noted that the siamese network architecture shares the parameters of GCN when training on pairs of input graphs $(G^1, G^2)$, and the number of GCN layers required depends on specific real-world applications.

\subsubsection{Node-Graph Matching Layer}\label{subsubsec:ngmn_matching}
This layer is the key part of our NGMN model, which can effectively learn the cross-level interactions by comparing each contextual node embedding of one graph against the whole graph-level embedding of another graph with the defined multi-perspective matching function.
There are generally two steps for this layer:
i) calculate the graph-level embedding vector of a graph;
ii) compare the node embeddings of a graph with the associated graph-level embedding vector of the other whole graph and then produce a similarity feature vector.

A simple approach to obtain the graph-level embedding vector of a graph is to directly perform pooling operations, \eg, element-wise max pooling.
However, this approach does not consider any information from the node embeddings that the resulting graph-level embedding vector will compare with later.
To build more tight and informative interactions between the two graphs for learning the graph-level embedding vector of each other, we first calculate a cross-graph attention coefficient $\alpha_{i,j}$ between the node $v_i \in \mathcal{V}^1$ in $G^1$ and all other nodes $v_j \in \mathcal{V}^2$ in $G^2$.
Similarly, we calculate the cross-graph attention coefficient $\beta_{j,i}$ between the node $v_j \in \mathcal{V}^2$ in $G^2$ and all other nodes $v_i \in \mathcal{V}^1$ in $G^1$.
To be specific, these two cross-graph attention coefficients can be computed with an attention function $f_s$ independently,
\begin{equation}
\begin{split}
    \alpha_{i,j} &= f_s(\vec{h}^1_i, \vec{h}^2_j) = \text{cosine} (\vec{h}^1_i, \vec{h}^2_j), \; v_j \in \mathcal{V}^2 \\
    \beta_{j,i} &= f_s(\vec{h}^2_j, \vec{h}^1_i) = \text{cosine} (\vec{h}^2_j, \vec{h}^1_i), \; v_i \in \mathcal{V}^1
\end{split}  
\end{equation}
where $f_s$ is the attention function for computing the similarity score between two node embedding vectors.
For simplicity, we use the cosine function in our experiments but other similarity metrics can be adopted as well.

Then, from the view of the node in one graph, we try to learn the corresponding attentive graph-level embedding vector of another graph.
Specifically, 
from the view of the node $v_i \in \mathcal{V}^1$ in $G^1$, we compute the attentive graph-level embedding vector $\widetilde{\vec{h}}^{2,i}_{G,avg}$ of $G^2$ by weighted averaging all node embeddings of $G^2$ with corresponding attentions.
Likewise, $\widetilde{\vec{h}}^{1,j}_{G,avg}$ is computed as the attentive graph-level embedding vector of $G^1$ from the view of the node $v_j \in \mathcal{V}^2$ in $G^2$.
Thus, we compute these two attentive graph-level embeddings as follows.
\begin{equation}\label{eq:attention weights}
\begin{split}
    \widetilde{\vec{h}}^{2,i}_{G,avg} &= \sum_{j \in \mathcal{V}^2} \alpha_{i,j} \vec{h}^2_j, \; v_i \in \mathcal{V}^1  \\
    \widetilde{\vec{h}}^{1,j}_{G,avg} &= \sum_{i \in \mathcal{V}^1} \beta_{j,i} \vec{h}^1_i, \; v_j \in \mathcal{V}^2  
\end{split}
\end{equation}

Next, we define a multi-perspective matching function $f_m$ to compute the similarity feature vector by comparing two input vectors of $\vec{x}_1$ and $\vec{x}_2$.
\begin{equation}\label{eq:multi-perspective defination}
\begin{split}
    \widetilde{\vec{h}}[k] &= f_m ( \vec{x}_1, \vec{x}_2, \vec{w}_k) \\
     &= \text{cosine} (\vec{x}_1 \odot \vec{w}_k, \vec{x}_2 \odot \vec{w}_k), \; k = {1,\dots,\widetilde{d}} 
\end{split}
\end{equation}
where 
$\odot$ is the element-wise multiplication operation,
$\widetilde{\vec{h}}[k] \in \mathcal{R}$ denotes the output similarity score in terms of $k$-th perspective, 
and $\vec{w}_k \in \mathcal{R}^{d'}$ represents the learnable weight vector in the $k$-th perspective.
When considering a total of $\widetilde{d}$ number of perspectives for the multi-perspective matching function $f_m$, the trainable weighted matrix will be $W_m = \{ \vec{w}_k\}_{k=1}^{\widetilde{d}} \in \mathcal{R}^{ d' \times \widetilde{d}}$.
After that, we will obtain a $\widetilde{d}$-dimension vector of similarity features, \ie, $\widetilde{\vec{h}} \in \mathcal{R}^{\widetilde{d}}$.

It is worth noting that the proposed $f_m$ essentially shares a similar spirit with the multi-head attention mechanism~\cite{vaswani2017attention}.
However, the most significant difference is that, the multi-head attention mechanism employs $\widetilde{d}$ number of trainable weighted \textit{matrices}, while $f_m$ uses $\widetilde{d}$ number of trainable weighted \textit{vectors} instead.
It is obvious that our methods use substantially fewer training parameters, which may reduce potential over-fitting as well as significantly speed up our computations.

With the defined multi-perspective matching function $f_m$ in Equation~(\ref{eq:multi-perspective defination}), we use it to compare the $i$-th node embedding in graph $G^1$ with the corresponding attentive graph-level embedding $\widetilde{\vec{h}}^{2,i}_{G,avg}$ of the other graph $G^2$.
The resulting similarity feature vector $\widetilde{\vec{h}}^1_i \in \mathcal{R}^{\widetilde{d}}$ is thus considered as the updated node embedding of $i$-th node in graph $G^1$.
Similarly, we also use $f_m$ to compare the $j$-th node embedding in the graph $G^2$ with the attentive graph-level embedding $\widetilde{\vec{h}}^{1,j}_{G,avg}$ of graph $G^1$, and consider the resulting feature vector $\widetilde{\vec{h}}^2_i \in \mathcal{R}^{\widetilde{d}}$ as the update node embedding of $j$-th node in the graph $G^2$.
Specifically, for all nodes in both $G^1$ and $G^2$, their corresponding node-graph interaction features can thus be computed by,
\begin{equation}\label{eq:multi-perspective operation}
\begin{split}
    \widetilde{\vec{h}}^1_i &= f_m ( \vec{h}^1_i, \widetilde{\vec{h}}^{2,i}_{G,avg}, W_m), \; v_i \in \mathcal{V}^1 \\
    \widetilde{\vec{h}}^2_j &= f_m ( \vec{h}^2_j, \widetilde{\vec{h}}^{1,j}_{G,avg}, W_m), \; v_j \in \mathcal{V}^2
\end{split}   
\end{equation}

After performing the above node-graph matching layer over all nodes for both graphs, these newly generated interaction features of nodes are considered and collected as the new feature matrices for $G^1$ and $G^2$,
\ie, $\widetilde{H}^1 = \{\widetilde{\vec{h}}^1_i\}_{i=1}^N \in \mathcal{R}^{N \times \widetilde{d}}$ and $\widetilde{H}^2 = \{\widetilde{\vec{h}}^2_j\}_{j=1}^M \in \mathcal{R}^{M \times \widetilde{d}}$,
which capture the cross-level interaction features between each node embeddings of one graph and a corresponding graph-level embedding of the other graph.

\subsubsection{Aggregation Layer}\label{subsubsec:ngmn_aggregation}
To aggregate the learned interactions from the node-graph matching layer, we employ the bidirectional LSTM (\ie, BiLSTM)~\cite{lstm:hochreiter1997long,melamud2016context2vec} to aggregate an unordered set of node embeddings for each graph as follows.

\begin{equation}\label{ngmn:aggregation}
    \widetilde{\vec{h}}_G^l =  \text{BiLSTM} \Big( \{\widetilde{\vec{h}}^l_i\}_{i=1}^{\{N,M\}} \Big) , \; l = \{1, 2\}
\end{equation}

Here, $\text{BiLSTM}$ in Equation~(\ref{ngmn:aggregation}) takes a random permutation of the node embeddings as the input and concatenates the two last hidden vectors from both directions (\ie, forward and backward) of the bidirectional LSTM as the representation of each graph.
The resulting $\widetilde{\vec{h}}_G^l \in \mathcal{R}^{2\widetilde{d}}$ represents the aggregated graph-level embedding vector for the graph $G^1$ or $G^2$.

Conceptually, an aggregator function should be invariant to permutations of its inputs while maintaining a large expressive capacity for aggregation.
However, we take BiLSTM as the default aggregation function, which is not permutation invariant on the set of node embeddings.
The reason is \textbf{three-fold} as follows.
(1) LSTM-related aggregators have been employed in previous work~\cite{GraphSage:hamilton2017inductive, HGNN:zhang2019heterogeneour} and have shown superior performance than other aggregators;
(2) To reduce the possible impact of the order of the input set on the BiLSTM aggregator, we take a random permutation of the set of node embeddings before inputting the Bidirectional LSTM model;
(3) We also conduct extensive experiments on the choice of aggregators in NGMN and show that BiLSTM achieves consistently better performance than other aggregator functions (\eg, see Table~\ref{tab:classification_all_baseline} and Table~\ref{tab:regression_all_baseline} in Section~\ref{subsec:effective} for details).
One possible reason we conjecture is that the BiLSTM aggregator offers a larger model capacity than other aggregators (\eg, Max/FCMax), potentially leading to more expressive ability for learning the graph-level embedding as an aggregation model.

\subsubsection{Prediction Layer}\label{subsubsec:ngmn_prediction}
After the aggregated graph-level embedding vectors $\widetilde{\vec{h}}_G^1$ and $\widetilde{\vec{h}}_G^2$ are obtained, we then use them to compute a similarity score $s(G^1, G^2)$ between $G^1$ and $G^2$.
Depending on the specific tasks, \ie, the graph-graph classification task and the graph-graph regression task, we have slightly different ways to calculate the final predicted similarity score as follows.

For the \textit{graph-graph classification} task, we directly compute the cosine similarity of two graph-level embedding vectors as follows, as it is quite common to employ the cosine similarity in other classification tasks~\cite{xu2017neural,gu2018deep,ling2021deep}.
\begin{equation}\label{eq:prediction:classification}
    \widetilde{y} = s(G^1, G^2) = \text{cosine} \Big( \widetilde{\vec{h}}_G^1, \widetilde{\vec{h}}_G^2 \Big)
\end{equation}

Differently, the predicted result of the graph-graph regression task is continuous and is normalized in a range of (0,1].
Thus, for the \textit{graph-graph regression} task, we first concatenate the two graph-level embedding vectors into $[\widetilde{\vec{h}}_G^1; \widetilde{\vec{h}}_G^2]$ and then employ four standard fully connected layers to gradually project the dimension of the resulting vector down to a scalar of the dimension 1.
Since the expected similarity score $\widetilde{y}$ should be in the range of (0, 1], we perform the \text{sigmoid} activation function to enforce the similarity score in this range.
We therefore compute the similarity score for the graph-graph regression task as follows.
\begin{equation}\label{eq:prediction:regression}
    \widetilde{y} = s(G^1, G^2) = \text{sigmoid} \Big(\text{MLP} \Big(\Big[\widetilde{\vec{h}}_G^1 ; \widetilde{\vec{h}}_G^2\Big] \Big)\Big)
\end{equation}
where $[\cdot\ ;\ \cdot]$ denotes the concatenation operation over two input vectors and $\text{MLP}$ denotes the employed four fully connected layers.

\subsubsection{Model Training}
The model is trained on a set of $n$ training triplets of two input graph-structured objects and a scalar output score $(G^1_1,G^2_1,y_1),...,(G^1_n,G^2_n,y_n) \in \mathcal{G} \times \mathcal{G} \times \mathcal{Y}$.
For both the graph-graph classification and graph-graph regression tasks, we train the models with the loss function of mean square error to compare the computed similarity score $\widetilde{y}$ with the ground-truth similarity score $y$.
\begin{equation}\label{eq:loss_function}
    \mathcal{L} = \frac{1}{n} \sum_{i=1}^{n} ( \widetilde{y} - y )^2
\end{equation}

\subsection{SGNN for Global-Level Interaction Learning}\label{subsec:sgnn}
The graph-level embeddings contain important information of a graph.
Therefore, learning graph-level interaction features between two graphs could be an important supplementary component for learning the graph similarity between two graphs.
In order to capture the global-level interaction features between two graphs, we present the siamese graph neural network (SGNN) which is also based on siamese network architecture as presented in the previous Section~\ref{subsubsec:ngmn_node_embedding}.
To be specific, our SGNN consists of three layers:
1) node embedding layer;
2) aggregation layer;
3) prediction layer.
We detail each layer of SGNN in the following.

\subsubsection{Node Embedding Layer}
For the sake of simplicity, we also adopt a multi-layer GCN with the siamese network architecture to generate contextual embeddings for all nodes, \ie, $H^1 = \{\vec{h}^1_i\}_{i=1}^N \in \mathcal{R}^{N \times d'}$ and $H^2 = \{\vec{h}^2_j\}_{j=1}^M \in \mathcal{R}^{M \times d'}$, for both graphs $G^1$ and $G^2$, respectively.
This is the same as the node embedding layer of NGMN that has already been explored in Section~\ref{subsubsec:ngmn_node_embedding}.

Conceptually, the node embedding layer in SGNN could be chosen to be independent or shared with the node embedding layer in NGMN.
As shown in Fig.~\ref{fig:hgmn_model}, our SGNN model shares the same node embedding layer with NGMN due to two reasons:
i) sharing the GCN parameters in the node embedding layer means reducing the number of parameters by half, which helps mitigate possible over-fitting;
ii) the shared GCN models maintain the consistency of the resulting node embeddings for both NGMN and SGNN, potentially leading to more aligned cross-level and graph-level interactions.
After all the node embeddings $H^1$ and $H^2$ for two graphs have been computed, they will be fed into the subsequent aggregation layer.

\subsubsection{Aggregation Layer}
With the computed node embeddings $H^1$ and $H^2$ for both $G^1$ and $G^2$, we need to aggregate them to formulate their corresponding graph-level embedding vectors $\vec{h}_G^1$ and $\vec{h}_G^2$ as follows.
\begin{equation}
    \vec{h}_G^l = \text{Aggregation} \Big( \{\vec{h}^l_i\}_{i=1}^{\{N,M\}} \Big) , \; l = \{1, 2\}
\end{equation}
where $\text{Aggregation}$ represents the aggregation function that outputs a corresponding graph-level embedding vector.
Without a doubt, we can use the BiLSTM aggregator function that has been introduced in Section~\ref{subsubsec:ngmn_aggregation}.
In addition to BiLSTM, we would like to employ other \textit{simpler} aggregator functions, such as element-wise max pooling (Max) and element-wise max pooling following a transformation by applying a standard fully connected layer (FCMax).

\subsubsection{Prediction Layer}
After the aggregated graph-level embedding vectors $\vec{h}_G^1$ and $\vec{h}_G^2$ are obtained, we then use the two vectors to compute the similarity score of $s(G^1, G^2)$.
Just like the prediction layer in NGMN, we use Equation~(\ref{eq:prediction:classification}) and Equation~(\ref{eq:prediction:regression}) to predict the similarity score for the graph-graph classification task and the graph-graph regression task, respectively.
We also use the same loss function of mean square error in Equation~(\ref{eq:loss_function}) to train SGNN.
In this way, we can also easily compare the performance difference between SGNN and NGMN.

\textbf{Compared with other graph-graph interactions.}
Although SimGNN~\cite{bai2019simgnn} also learns graph-graph interaction features, our SGNN is still different from it in three aspects.
i) We apply a siamese network architecture with one shared multi-layer GCN model to learn node embeddings rather than two independent multi-layer GCN models;
ii) SGNN only employs a simple aggregation function to learn a graph-level embedding vector while SimGNN uses a context-aware attention method;
iii) SGNN directly employs the cosine function or the concatenation with fully connected layers to learn the graph-graph interaction features, whereas SimGNN uses a more sophisticated neural tensor network~\cite{socher2013reasoning} to capture the graph-graph interaction features.

\subsection{Discussions on Our Full Model -- \system}
\subsubsection{\system}
The full model \system combines the advantages of both NGMN and SGNN to capture both the cross-level node-graph interactions and global-level graph-graph interactions for better representation learning in computing the graph similarity between two graphs.
As shown in Fig.~\ref{fig:hgmn_model}, for the full model \system, the two input graphs $G^l \, (l = \{1, 2\})$ first go through NGMN and SGNN, which produce two corresponding graph-level embedding vectors, \ie, $\widetilde{\vec{h}}_G^l$ and $\vec{h}_G^l$ with $l = \{1, 2\}$, respectively.
After that, we concatenate the two aggregated graph-level embedding vectors from NGMN and SGNN for each graph, and then feed those concatenated embeddings into the following prediction layer as presented in Section~\ref{subsubsec:ngmn_prediction}.
Particularly, we compute the similarity score $s(G^1, G^2)$ between $G^1$ and $G^2$ for both the graph-graph classification and graph-graph regression tasks as follows.
\begin{equation}
 \widetilde{y} = s(G^1, G^2) = \text{cosine} ( \big[ \widetilde{\vec{h}}_G^1 ; \vec{h}_G^1 \big], \big[ \widetilde{\vec{h}}_G^2 ; \vec{h}_G^2 \big])
\end{equation}
\begin{equation}
\resizebox{.89\linewidth}{!}{$
 \widetilde{y} = s(G^1, G^2) = \text{sigmoid} \Big(\text{MLP} \Big(\Big[\big[\widetilde{\vec{h}}_G^1 ; \vec{h}_G^1 \big] ; \big[\widetilde{\vec{h}}_G^2 ; \vec{h}_G^2 \big]\Big] \Big)\Big) 
$}
\end{equation}

\subsubsection{Complexity Analysis}
As the dominant computation of SGNN is to generate node embeddings for both graphs by Equation~(\ref{eq:gcn}), its time complexity is $\mathcal{O}(max(|\mathcal{E}^1|, |\mathcal{E}^2|)$, in which $|\mathcal{E}^1|$ and $|\mathcal{E}^2|$ denote the number of edges in both graphs, respectively.
Similarly, the most computationally extensive operations of NGMN are in Equation (\ref{eq:attention weights}) and Equation (\ref{eq:multi-perspective operation}), the time complexity of NGMN is $\mathcal{O}(NM)$, where $N$ and $M$ are the number of nodes in both graphs, respectively.
For the full model \system, the overall time complexity is thus $\mathcal{O}(max(|\mathcal{E}^1|, |\mathcal{E}^2|) + \mathcal{O}(NM) = \mathcal{O}(NM)$.
Despite \system has a time complexity of $\mathcal{O}(NM)$, we argue that it is still acceptable due to the following three reasons:
\begin{enumerate}
    \item As the GED computation is NP-hard, the time complexity of exact solutions like $A^{*}$~\cite{hart1968formal} is normally $\mathcal{O}(N^M)$.
    Although recent approximations have reduced to $\mathcal{O}((N+M)^3)$ or even $\mathcal{O}((N+M)^2)$~\cite{fischer2015approximation}, \system still shows much lower time complexity;
    
    \item Compared with state-of-the-art baseline models~\cite{bai2019simgnn,bai2020learning,li2019graph} that have a time complexity of $\mathcal{O}(max(N, M)^{2})$, \system does not increase the time complexity.
    
    \item As the entire model architecture of \system is built on GNNs, all matrix multiplications can be greatly accelerated with GPUs or distributed data-parallel training.
\end{enumerate}
\section{Evaluation}\label{sec:experiment}
In this section, we systematically evaluate the performance of our full model \system compared with recently proposed baseline models on four benchmark datasets for both the graph-graph classification and graph-graph regression tasks.
In particular,
we first introduce the benchmark datasets of both tasks in Section~\ref{subsec:datasets},
provide details of the experimental settings in Section~\ref{subsec:settings},
describe the baseline models to be compared with our models in Section~\ref{subsec:baseline},
and finally present the main evaluation results as well as conduct ablation studies in Section~\ref{subsec:effective} and Section~\ref{subsec:ablation}, respectively.

\begin{table*}[ht]
  \centering
  \caption{Summary statistics of datasets for both the graph-graph classification task \& the graph-graph regression task.}
  \label{table:datasetstatistics}
  \renewcommand\tabcolsep{5.2pt} 
  \begin{tabular}{ccrrrrrc}
    \toprule
    Tasks & Datasets & \tabincell{c}{Sub-Datasets} & \tabincell{c}{\# of Graphs} & \tabincell{c}{\tabincell{c}{\# of Functions}} & \tabincell{c}{AVG \# of Nodes} & \tabincell{c}{AVG \# of Edges}  & \tabincell{c}{Initial Feature Dimensions} \\
    \midrule
    \multicolumn{1}{c}{\multirow{6}{*}{\tabincell{c}{Graph-Graph\\Classification Task}}}
    & \multirow{3}{*}{\bf FFmpeg}   &   [3, 200]    & 83,008    & {10,376}      & 18.83     & 27.02     & \multirow{3}{*}{6}    \\
    & \multicolumn{1}{c}{}          &   [20, 200]   & 31,696    & {7,668}       & 51.02     & 75.88     &                       \\
    & \multicolumn{1}{c}{}          &   [50, 200]   & 10,824    & {3,178}       & 90.93     & 136.83    &                       \\
    \cmidrule{2-8}
    & \multirow{3}{*}{\bf OpenSSL}  &    [3, 200]   & 73,953    & {4,249}       & 15.73     & 21.97     & \multirow{3}{*}{6}    \\
    & \multicolumn{1}{c}{}          &   [20, 200]   & 15,800    & {1,073}       & 44.89     & 67.15     &                       \\
    & \multicolumn{1}{c}{}          &   [50, 200]   & 4,308     & {338}         & 83.68     & 127.75    &                       \\
    \midrule
    \multicolumn{1}{c}{\multirow{2}{*}{\tabincell{c}{Graph-Graph\\Regression Task}}}
    & \bf AIDS700                   &  -            & 700       & -             & 8.90      & 8.80      & 29                    \\
    \cmidrule{2-8}
    & \bf LINUX1000                 & -             & 1,000     & -             & 7.58      & 6.94      & 1                     \\
    \bottomrule
  \end{tabular}
\end{table*}

\subsection{Tasks \& Datasets}\label{subsec:datasets}
\subsubsection{Graph-Graph Classification Task \& Datasets}
We evaluate our models on the graph-graph classification task of computing a similarity score (\ie, $y \in \{-1, 1\}$) between two binary functions, which is the heart of many binary security problems~\cite{xu2017neural, ding2019asm2vec, li2019graph}.
Conceptually, two binary functions that are compiled from the same source code but under different settings (\eg, architectures, compilers, optimization levels, etc) are considered to be semantically similar to each other.
To learn the similarity score of a pair of binary functions, we represent each binary function with a control flow graph (CFG), whose nodes represent the basic blocks (a basic block is a sequence of instructions without jumps) and edges represent control flow paths between these basic blocks.
Thus, computing a similarity score between two binary functions can be cast as the problem of learning the similarity score $s(G^1, G^2)$ between two CFGs $G^1$ and $G^2$,
where $s(G^1,G^2) = +1$ indicates $G^1$ and $G^2$ are similar; otherwise $s(G^1,G^2) = -1$ indicates dissimilar.
We prepare two benchmark datasets generated from two popular open-source software: \textbf{FFmpeg}\footnote{\url{https://www.ffmpeg.org/}} and \textbf{OpenSSL}\footnote{\url{https://www.openssl.org/}}, to evaluate our models on the graph-graph classification task.
More details about how we prepare both datasets can be found in Appendix~\ref{appendix:more_datasets}.

Besides, existing work does not consider the impact of the graph size on the performance of graph matching networks.
However, we find the larger the graph size is, the worse the performance is.
Therefore, it is important to evaluate the robustness of graph matching models under this setting.
We thus further split each dataset into 3 sub-datasets (\ie, [3, 200], [20,200], and [50,200]) according to the size ranges of pairs of input graphs.

It is noted that, although there are many benchmark datasets~\cite{hu2020open} for the general graph classification task~\cite{ying2018hierarchical,ma2019graph}, these datasets cannot be directly employed in our graph-graph classification task as we cannot treat the two input graphs with the same labels as ``similar''.
This is because the general graph classification task only assigns each graph with a label, while our graph-graph classification task learns a binary similarity label (\ie, similar or dissimilar) for pairs of two graphs instead of one graph.

\subsubsection{Graph-Graph Regression Task \& Datasets}
We evaluate our models on learning the graph edit distance (GED)~\cite{gao2010survey, riesen2015structural}, which measures the structural similarity between two input graphs.
Formally, GED is defined as the cost of the least sequence of edits that transforms one graph into another graph, where an edit can be an insertion/deletion of a node/edge.
Instead of directly computing the GED between two $G^1$ and $G^2$, we try to learn a similarity score $s(G^1, G^2)$, which is the normalized exponential of GED, \ie, $exp^{-\big[\frac{GED(G^1,G^2)}{(N+M)/2}\big]}$ in the range of $(0, 1]$.
In particular, we employ two benchmark datasets, \ie, \textbf{AIDS700} and \textbf{LINUX1000}, which are released by~\cite{bai2019simgnn} and publicly available.\footnote{\url{https://github.com/yunshengb/SimGNN}}
Each dataset contains a set of pairs of input graphs as well as their ground-truth GED scores, which are computed by exponential-time exact GED computation algorithm $A^{*}$~\cite{hart1968formal,riesen2013novel}.
As the ground-truth GEDs of another dataset {IMDB-MULTI} are provided with \textit{in-exact} approximations, we thus do not consider this dataset in our evaluation.

In summary, for both the graph-graph classification and graph-graph regression tasks, 
we follow the same training/validation/testing split as previous work~\cite{bai2019simgnn, li2019graph} for fair comparisons.
For the graph-graph classification task, {Bai et al.} split each dataset into three disjoint subsets of binary functions in which 80\% for training, 10\% for validation and 10\% for testing.
For each dataset in the graph-graph regression task, {Li et al.} randomly split 60\%, 20\%, and 20\% of all graphs as the training/validation/testing subsets, and then build the pairwise datasets.
Table~\ref{table:datasetstatistics} shows the summary statistic of all datasets.

\subsection{Implementation Settings}\label{subsec:settings}
To set up our models, including SGNN, NGMN, and \system, we use a three-layer GCN in the node embedding layer and each of the GCN layers has an output dimension of 100.
We use ReLU as the activation function along with a dropout layer after each GCN layer with the dropout rate being 0.1.
In the aggregation layer of SGNN, we can employ different aggregation functions (\ie, Max, FCMax, and BiLSTM) as introduced previously in Section~\ref{subsec:sgnn}.
For NGMN, we set the number of perspectives $\widetilde{d}$ to 100.
For NGMN, we take BiLSTM as the default aggregation function and we make its hidden size equal to the dimension of node embeddings.
For each graph, we concatenate the two last hidden vectors of both directions of BiLSTM, which results in a 200-dimension vector as the graph-level embedding vector.

Our implementation is built using PyTorch~\cite{paszke2019pytorch} and PyTorch Geometric~\cite{Fey2019PytorchGeometric}.
To train our models, we use the Adam optimizer~\cite{adam:kingma2014adam}.
For the graph-graph classification task, we train the model by running 100 epochs with 0.5e-3 learning rate.
At each epoch, we build the pairwise training data as follows.
For each graph $G$ in the training set, we obtain one positive pair $\{(G, G^{pos}), +1\}$ and a corresponding negative pair $\{(G, G^{neg}), -1\}$, where $G^{pos}$ is randomly selected from all CFGs that are compiled from the same source function as $G$, and $G^{neg}$ is selected from the other graphs.
By default, each batch includes 5 positive and 5 negative pairs.
For the graph-graph regression task,
we train the model by running 10,000 iterations with a batch of 128 graph pairs with 5e-3 learning rate.
Each pair is a tuple of $\{(G^{1}, G^{2}), s\}$, where $s$ is the ground-truth normalized GED between $G^{1}$ and $G^{2}$.
It is noted that all experiments are conducted on a PC equipped with 8 Intel Xeon 2.2GHz CPU and one NVIDIA GTX 1080 Ti GPU.

\subsection{Baseline Methods}\label{subsec:baseline}
To evaluate the effectiveness of our models, we consider 3 state-of-the-art baseline models for comparisons as follows.
\begin{itemize}
    \item [1)]
    \textit{SimGNN}~\cite{bai2019simgnn} adopts a multi-layer GCN and employs two strategies to calculate the GED between two graphs: one uses a neural tensor network to capture the graph-graph interactions; another uses histogram features extracted from two sets of node embeddings;
    \item [2)]
    \textit{GMN}~\cite{li2019graph} employs a variant of message passing neural networks and improves the node embeddings of one graph via incorporating the information of attentive neighborhoods of another graph;
    \item [3)]
    \textit{GraphSim}~\cite{bai2020learning} extends SimGNN by first turning the two sets of node embeddings into a similarity matrix and then processing the matrix with CNNs.
\end{itemize}

As the three baselines only consider either the graph-graph classification task or the graph-graph regression task,
we slightly adjust the last layer of the model or the loss function of each baseline to make a fair comparison on both tasks.
Detailed experimental settings of these baselines are given in Appendix~\ref{appendix:baseline_settings}.
It is also worth noting that we repeat all experiments 5 times and report the mean and standard deviation of experimental results, with the best in \textbf{bold}.
\subsection{Evaluation Results the Proposed Models}\label{subsec:effective}
In this subsection, we compare our proposed models with all three state-of-the-art baseline models on both the graph-graph classification and graph-graph regression tasks.
Specifically, our models contain a full model \bfsystem and two partial models, \ie, SGNN and NGMN.
For both SGNN and NGMN, we employ three different aggregator functions (\ie, Max, FCMax, and BiLSTM) in their corresponding aggregation layers.
As the full model \bfsystem combines both SGNN and NGMN, \bfsystem must have two aggregation functions for SGNN and NGMN.
In the following subsection, we use the bracket after the model indicates the employed aggregator function(s).
For instance, SGNN (Max) represents the SGNN model with the Max aggregator and NGMN (BiLSTM) denotes the NGMN model with the BiLSTM aggregator;
\bfsystem (FCMax + BiLSTM) means that \bfsystem takes FCMax as the aggregator for SGNN and BiLSTM as the aggregator for NGMN.

\subsubsection{Graph-Graph Classification Task}
For the graph-graph classification task of detecting whether two binary functions are similar or not,
we measure the Area Under the ROC Curve (AUC)~\cite{auc:bradley1997the} for classifying pairs of input graphs in the same testing dataset.
AUC is one of the most commonly used evaluation metrics to evaluate the binary classification models as AUC is independent of the manually selected threshold.
Apparently, the larger AUC, the better performance of the models.
The main results are shown in Table~\ref{tab:classification_all_baseline}.

We first investigate the impact of different aggregation functions employed by our models, including SGNN, NGMN, and \bfsystem.
For SGNN and NGMN, it is clearly seen from
Table~\ref{tab:classification_all_baseline} that the Max aggregator achieves the worst performance and the BiLSTM aggregator offers better performance on both \textbf{FFmpeg} and \textbf{OpenSSL} datasets.
One possible reason we conjecture is that BiLSTM might admit more expressive ability as the aggregation model.
In addition, when comparing the results of NGMN (BiLSTM) with SGNN (BiLSTM), NGMN (BiLSTM) offers superior performance on almost all sub-datasets, demonstrating the benefits of the node-graph matching mechanism that captures the cross-level interactions between two graphs.

As NGMN (BiLSTM) shows the best performance, for the full model \bfsystem, we fix BiLSTM as the default aggregator for the NGMN part and vary the aggregator (\eg, Max, FCMax, or BiLSTM) for the SGNN part.
Therefore, compared \bfsystem with the two partial models (\ie, SGNN and NGMN),
it is observed that \bfsystem further improves the performance of NGMN together with the graph-level interaction features learned from SGNN.
This confirms that SGNN and NGMN are two indispensable partial models for our full model \bfsystem.

Besides, we also compare our full model \bfsystem with 3 baseline models.
The experimental results that our models with the BiLSTM aggregator clearly achieve the state-of-the-art performance on all 6 sub-datasets for both \textbf{FFmpeg} and \textbf{OpenSSL} datasets.
Particularly when the graph size increases, both \bfsystem and the key component NGMN (BiLSTM) models show better performance and robustness than state-of-the-art methods.

\begin{table*}[ht]
    \caption{Summary of experimental results on the graph-graph classification task in terms of AUC scores (\%).}
    \label{tab:classification_all_baseline}
    \renewcommand\tabcolsep{12pt} 
    \centering
      \begin{tabular}{lcccccc}
      \toprule
      \multicolumn{1}{c}{\multirow{3}{*}{Model}} & \multicolumn{3}{c}{\bf FFmpeg} & \multicolumn{3}{c}{\bf OpenSSL} \\
      \cmidrule(lr){2-4} \cmidrule(lr){5-7}
      \multicolumn{1}{c}{} & \multicolumn{1}{c}{[3, 200]} & \multicolumn{1}{c}{[20, 200]} & \multicolumn{1}{c}{[50, 200]} & \multicolumn{1}{c}{[3, 200]} & \multicolumn{1}{c}{[20, 200]} & \multicolumn{1}{c}{[50, 200]} \\
    \midrule
    \midrule
      SimGNN                        & 95.38$\pm$0.76     & 94.31$\pm$1.01     & 93.45$\pm$0.54     & 95.96$\pm$0.31     & 93.58$\pm$0.82     & 94.25$\pm$0.85 \\
      GMN                           & 94.15$\pm$0.62     & 95.92$\pm$1.38     & 94.76$\pm$0.45     & 96.43$\pm$0.61     & 93.03$\pm$3.81     & 93.91$\pm$1.65 \\
      GraphSim                      & 97.46$\pm$0.30     & 96.49$\pm$0.28     & 94.48$\pm$0.73     & 96.84$\pm$0.54     & 94.97$\pm$0.98     & 93.66$\pm$1.84 \\
    \midrule
      SGNN (Max)                    & 93.92$\pm$0.07     & 93.82$\pm$0.28     & 85.15$\pm$1.39     & 91.07$\pm$0.10     & 88.94$\pm$0.47     & 82.10$\pm$0.51 \\
      SGNN (FCMax)                  & 95.37$\pm$0.04     & 96.29$\pm$0.14     & 95.98$\pm$0.32     & 92.64$\pm$0.15     & 93.79$\pm$0.17     & 93.21$\pm$0.82 \\
      SGNN (BiLSTM)                 & 96.92$\pm$0.13     & 97.62$\pm$0.13     & 96.35$\pm$0.33     & 95.24$\pm$0.06     & 96.30$\pm$0.27     & 93.99$\pm$0.62 \\
    \midrule
      NGMN (Max)                    & 73.74$\pm$8.30     & 73.85$\pm$1.76     & 77.72$\pm$2.07     & 67.14$\pm$2.70     & 63.31$\pm$3.29     & 63.02$\pm$2.77 \\
      NGMN (FCMax)                  & 97.28$\pm$0.08     & 96.61$\pm$0.17     & 96.65$\pm$0.30     & 95.37$\pm$0.19     & 96.08$\pm$0.48     & \bf 95.90$\pm$0.73 \\
      NGMN (BiLSTM)                 & 97.73$\pm$0.11     & 98.29$\pm$0.21     & 96.81$\pm$0.96     & 96.56$\pm$0.12     & \bf 97.60$\pm$0.29     & 92.89$\pm$1.31 \\
    \midrule
    \bfsystem (Max + BiLSTM)        & 97.44$\pm$0.32     & 97.84$\pm$0.40     & 97.22$\pm$0.36     & 94.77$\pm$1.80     & 97.44$\pm$0.26     & 94.06$\pm$1.60 \\
    \bfsystem (FCMax + BiLSTM)      & \bf98.07$\pm$0.06  & \bf98.29$\pm$0.10  & \bf97.83$\pm$0.11  & 96.87$\pm$0.24     & 97.59$\pm$0.24     & 95.58$\pm$1.13 \\
    \bfsystem (BiLSTM + BiLSTM)     & 97.56$\pm$0.38     & 98.12$\pm$0.04     & 97.16$\pm$0.53     & \bf96.90$\pm$0.10  & 97.31$\pm$1.07     & 95.87$\pm$0.88 \\
    \bottomrule
    \end{tabular}
  \end{table*}
\begin{table*}[htb]
    \caption{Summary of experimental results on the graph-graph regression task in terms of $mse$, $\rho$, $\tau$, $p@10$ \& $p@20$.}
    \label{tab:regression_all_baseline}
    \renewcommand\tabcolsep{10pt} 
    \centering
      \begin{tabular}{clccccc}
      \toprule
      Datasets & \multicolumn{1}{c}{Model} & $mse$ ($10^{-3}$) & $\rho$ & $\tau$ & $p@10$ & $p@20$ \\
      \midrule
      \midrule
      \multicolumn{1}{c}{\multirow{14}{*}{\textbf{AIDS700}}}
      & SimGNN                              & 1.376$\pm$0.066     & 0.824$\pm$0.009    & 0.665$\pm$0.011      & 0.400$\pm$0.023    & 0.489$\pm$0.024    \\
      & GMN                                 & 4.610$\pm$0.365     & 0.672$\pm$0.036    & 0.497$\pm$0.032      & 0.200$\pm$0.018    & 0.263$\pm$0.018    \\
      & GraphSim                            & 1.919$\pm$0.060     & 0.849$\pm$0.008    & 0.693$\pm$0.010 & 0.446$\pm$0.027 & 0.525$\pm$0.021 \\
      \cmidrule{2-7} 
      & SGNN (Max)                          & 2.822$\pm$0.149      & 0.765$\pm$0.005      & 0.588$\pm$0.004      & 0.289$\pm$0.016      & 0.373$\pm$0.012 \\
      & SGNN (FCMax)                        & 3.114$\pm$0.114      & 0.735$\pm$0.009      & 0.554$\pm$0.008      & 0.278$\pm$0.021      & 0.364$\pm$0.017 \\
      & SGNN (BiLSTM)                       & 1.422$\pm$0.044      & 0.881$\pm$0.005      & 0.718$\pm$0.006      & 0.376$\pm$0.020      & 0.472$\pm$0.014 \\
      \cmidrule{2-7}    
      & NGMN (Max)                          & 2.378$\pm$0.244      & 0.813$\pm$0.015      & 0.642$\pm$0.013      & \bf0.578$\pm$0.199   & \bf0.583$\pm$0.169 \\
      & NGMN (FCMax)                        & 2.220$\pm$1.547      & 0.808$\pm$0.145      & 0.656$\pm$0.122      & 0.425$\pm$0.078      & 0.504$\pm$0.064 \\
      & NGMN (BiLSTM)                       & 1.191$\pm$0.048      & 0.904$\pm$0.003      & 0.749$\pm$0.005      & 0.465$\pm$0.011      & 0.538$\pm$0.007 \\
      \cmidrule{2-7}
      & \bfsystem (Max + BiLSTM)            & 1.210$\pm$0.020      & 0.900$\pm$0.002      & 0.743$\pm$0.003      & 0.461$\pm$0.012      & 0.534$\pm$0.009      \\
      & \bfsystem (FCMax + BiLSTM)          & 1.205$\pm$0.039      & 0.904$\pm$0.002      & 0.749$\pm$0.003      & 0.457$\pm$0.014      & 0.532$\pm$0.016      \\
      & \bfsystem (BiLSTM + BiLSTM)         & \bf1.169$\pm$0.036   & \bf0.905$\pm$0.002   & \bf0.751$\pm$0.003   & 0.456$\pm$0.019      & 0.539$\pm$0.018   \\
      \midrule
      \midrule
      \multicolumn{1}{c}{\multirow{14}{*}{\textbf{\tabincell{c}{LINUX1000}}}}
      & SimGNN                              & 2.479$\pm$1.038       & 0.912$\pm$0.031      & 0.791$\pm$0.046      & 0.635$\pm$0.328      & 0.650$\pm$0.283 \\
      & GMN                                 & 2.571$\pm$0.519       & 0.906$\pm$0.023      & 0.763$\pm$0.035      & 0.888$\pm$0.036      & 0.856$\pm$0.040 \\
      & GraphSim                            & 0.471$\pm$0.043       & 0.976$\pm$0.001      & \bf{0.931$\pm$0.003} & \bf{0.956$\pm$0.006} & 0.942$\pm$0.007 \\
      \cmidrule{2-7}
      & SGNN (Max)                          & 11.832$\pm$0.698     & 0.566$\pm$0.022      & 0.404$\pm$0.017      & 0.226$\pm$0.106      & 0.492$\pm$0.190 \\
      & SGNN (FCMax)                        & 17.795$\pm$0.406     & 0.362$\pm$0.021      & 0.252$\pm$0.015      & 0.239$\pm$0.000      & 0.241$\pm$0.000 \\
      & SGNN (BiLSTM)                       & 2.140$\pm$1.668      & 0.935$\pm$0.050      & 0.825$\pm$0.100      & 0.878$\pm$0.012      & 0.865$\pm$0.007 \\
      \cmidrule{2-7}
      & NGMN (Max)$^*$       & 16.921$\pm$0.000 & - & - & - & - \\
      & NGMN (FCMax)                        & 4.793$\pm$0.262      & 0.829$\pm$0.006      & 0.665$\pm$0.011      & 0.764$\pm$0.170      & 0.767$\pm$0.166 \\
      & NGMN (BiLSTM)                       & 1.561$\pm$0.020      & 0.945$\pm$0.002      & 0.814$\pm$0.003      & 0.743$\pm$0.085      & 0.741$\pm$0.086  \\
      \cmidrule{2-7}
      & \bfsystem (Max + BiLSTM)            & 1.054$\pm$0.086      & 0.962$\pm$0.003      & 0.850$\pm$0.008      & 0.877$\pm$0.054      & 0.883$\pm$0.047      \\
      & \bfsystem (FCMax + BiLSTM)          & 1.575$\pm$0.627      & 0.946$\pm$0.019      & 0.817$\pm$0.034      & 0.807$\pm$0.117      & 0.784$\pm$0.108      \\
      & \bfsystem (BiLSTM + BiLSTM)         & \bf{0.439$\pm$0.143} & \bf0.985$\pm$0.005   & 0.919$\pm$0.016      & 0.955$\pm$0.011      & \bf0.943$\pm$0.014   \\
      \bottomrule
      \end{tabular}
     \footnotesize{\\$^*$ As all duplicated experiments running on this setting do not converge in their training processes, their corresponding results cannot be calculated.}
\end{table*}

\subsubsection{Graph-Graph Regression Task}
For the graph-graph regression task of computing the similarity score GED between two input graphs, we measure the Mean Square Error ($mse$), Spearman’s Rank Correlation Coefficient ($\rho$)~\cite{pho:spearman1904the}, Kendall’s Rank Correlation Coefficient ($\tau$)~\cite{tau:kendall1938new}, and precision at k ($p@k$) as previous work~\cite{bai2019simgnn,bai2020learning} for fair comparisons.
Apparently, the smaller $mse$, the better performance of models.
But for $\rho$, $\tau$, and $p@k$, the larger the better.
All results of both datasets are summarized in Table~\ref{tab:regression_all_baseline}.

For the impact of different aggregator functions employed by both SGNN and NGMN, our experimental results draw similar conclusions that the BiLSTM aggregator offers superior performance on both datasets and NGMN (BiLSTM) achieves better performance than SGNN (BiLSTM) in terms of most evaluation metrics.
Furthermore, compared \bfsystem with the two partial models (\ie, SGNN and NGMN), it is clearly seen that \bfsystem further improves the performance of either SGNN or NGMN.
These observations confirm again that NGMN and SGNN are two indispensable parts for the full model \bfsystem, which could capture both the cross-level node-graph interactions and global-level graph-graph interactions for better representation learning in computing the graph similarity between two graphs.

We also compare our full model \bfsystem with state-of-the-art baseline models on the graph-graph regression task.
From Table~\ref{tab:regression_all_baseline},
it can be observed that,
although GraphSim shows better performance than the other two baselines, our full model \bfsystem and its key component NGMN outperform all baselines on both \textbf{AIDS700} and \textbf{LINUX1000} datasets in terms of most evaluation metrics.

\textbf{Comparison of both tasks.}
When comparing the graph-graph regression task with the graph-graph classification task, it can be observed from Table~\ref{tab:classification_all_baseline} and Table~\ref{tab:regression_all_baseline} that, \system outperforms both SGNN and NGMN on the graph-graph regression task by quite a larger margin, but a lower margin on the graph-graph classification task.
One possible reason we conjecture is that different interactions (\ie, global-level, cross-level, or both) learned from different models (SGNN, NGMN, or MGMN) might have different impacts on different tasks.
For the graph-graph classification task in Table~\ref{tab:classification_all_baseline}, either SGNN or NGMN can easily achieve high performance.
Naturally, it makes MGMN that combines both SGNN and NGMN appear to be limited improved.
For the graph-graph regression task in Table~\ref{tab:regression_all_baseline}, both SGNN and NGMN offer limited performance.
However, when we combine both SGNN and NGMN into MGMN, it can further improve the overall performance in terms of all metrics by a larger margin.

\subsection{Ablation Studies on NGMN}\label{subsec:ablation}
In this subsection, we explore ablation studies to measure the contributions of different components in NGMN, which is the key partial model of the full model \bfsystem.
The reasons why we conduct ablation studies on NGMN instead of \bfsystem are as follows.
First, we have already shown the performance of SGNN, NGMN, and \bfsystem in the above Section~\ref{subsec:effective}, highlighting the importance of our proposed cross-level interactions learned from NGMN and graph-level interactions learned from SGNN.
Second, conducting only ablation studies on NGMN would make the evaluation results easier to be observed and clearer to be compared, as it excludes the potential influences of SGNN.

\subsubsection{Impact of Different Attention Functions}
As explored in Section~\ref{subsubsec:ngmn_matching}, the proposed multi-perspective matching function shares similar spirits with the multi-head attention mechanism~\cite{vaswani2017attention}, which makes it interesting to compare them.
Therefore, we investigate the impact of these two different mechanisms for NGMN with classification results shown in Table~\ref{tab:ablation:classification_multihead}.
In our evaluation, the number of heads $K$ in the multi-head attention model is set to 6 because of the substantial consumption of resources of multi-head attention computations.
Even though, the number of parameters of multi-head attention is still multiple times more than our multi-perspective matching function.
Interestingly, it is observed that our proposed multi-perspective matching function consistently outperforms the results of the multi-head attention by quite a large margin.
We suspect that the multi-perspective matching function uses attention weighted vectors rather than matrices, which may reduce the potential over-fitting.
\begin{table*}[htbp]
  \centering
  \caption{The graph-graph classification results of multi-perspectives versus multi-heads in terms of AUC scores (\%).}
  \renewcommand\tabcolsep{13.5pt}
  \label{tab:ablation:classification_multihead}
    \begin{tabular}{ccccccc}
    \toprule
    \multirow{3}{*}{Model} & \multicolumn{3}{c}{\bf FFmpeg}& \multicolumn{3}{c}{\bf OpenSSL} \\
    \cmidrule(lr){2-4} \cmidrule(lr){5-7}
    \multicolumn{1}{c}{} & \multicolumn{1}{c}{[3, 200]} & \multicolumn{1}{c}{[20, 200]} & \multicolumn{1}{c}{[50, 200]} & \multicolumn{1}{c}{ [3, 200]} & \multicolumn{1}{c}{[20, 200]} & \multicolumn{1}{c}{[50, 200]} \\
    \midrule
    Multi-Perspectives ($\widetilde{d}=100$) & \bf 97.73$\rpm$0.11 & \bf 98.29$\rpm$0.21 & \bf 96.81$\rpm$0.96 & \bf 96.56$\rpm$0.12 & \bf 97.60$\rpm$0.29 & \bf 92.89$\rpm$1.31 \\
    Multi-Heads ($\text{\# of Heads} = 6$)   & 91.18$\rpm$5.91	& 77.49$\rpm$5.21	& 68.15$\rpm$6.97	& 92.81$\rpm$5.21	& 85.43$\rpm$5.76 & 56.87$\rpm$7.53 \\
    \bottomrule
    \end{tabular}
\end{table*}
\begin{table*}[htbp]
  \centering
  \caption{The graph-graph classification results of NGMN models with different numbers of perspectives in terms of AUC scores (\%).}
  \label{tab:ablation:classification_different_perspectives}
  \renewcommand\tabcolsep{16.2pt}
    \begin{tabular}{lcccccc}
    \toprule
    \multicolumn{1}{c}{\multirow{3}{*}{Model}} & \multicolumn{3}{c}{\bf FFmpeg}& \multicolumn{3}{c}{\bf OpenSSL} \\
    \cmidrule(lr){2-4} \cmidrule(lr){5-7}
    \multicolumn{1}{c}{} & \multicolumn{1}{c}{[3, 200]} & \multicolumn{1}{c}{[20, 200]} & \multicolumn{1}{c}{[50, 200]} & \multicolumn{1}{c}{[3, 200]} & \multicolumn{1}{c}{[20, 200]} & \multicolumn{1}{c}{[50, 200]} \\
    \midrule
    NGMN-($\widetilde{d}=~50$)     & 98.11$\rpm$0.14   & 97.76$\rpm$0.14 & 96.93$\rpm$0.52 & \bf97.38$\rpm$0.11 & 97.03$\rpm$0.84 & 93.38$\rpm$3.03 \\
    NGMN-($\widetilde{d}=~75$)     & 97.99$\rpm$0.09   & 97.94$\rpm$0.14 & 97.41$\rpm$0.05 & 97.09$\rpm$0.25 & 98.66$\rpm$0.11 & 92.10$\rpm$4.37 \\
    NGMN-($\widetilde{d}=100$)    & 97.73$\rpm$0.11   & \bf98.29$\rpm$0.21 & 96.81$\rpm$0.96 & 96.56$\rpm$0.12 & 97.60$\rpm$0.29 & 92.89$\rpm$1.31 \\
    NGMN-($\widetilde{d}=125$)    & 98.10$\rpm$0.03   & 98.06$\rpm$0.08 & 97.26$\rpm$0.36 & 96.73$\rpm$0.33 & \bf98.67$\rpm$0.11 & 96.03$\rpm$2.08 \\
    NGMN-($\widetilde{d}=150$)    & \bf98.32$\rpm$0.05 & 98.11$\rpm$0.07 & \bf97.92$\rpm$0.09 & 96.50$\rpm$0.31 & 98.04$\rpm$0.03 & \bf97.13$\rpm$0.36 \\
    \bottomrule
    \end{tabular}
\end{table*}
\begin{table*}[htbp]
  \centering
  \caption{The graph-graph classification results of NGMN models with different GNNs in terms of AUC scores (\%).}
  \label{tab:ablation:classification_othergnn}
  \renewcommand\tabcolsep{15.6pt}
    \begin{tabular}{lcccccc}
    \toprule
    \multicolumn{1}{c}{\multirow{3}{*}{Model}} & \multicolumn{3}{c}{\bf FFmpeg}& \multicolumn{3}{c}{\bf OpenSSL} \\
    \cmidrule(lr){2-4} \cmidrule(lr){5-7}
    \multicolumn{1}{c}{} & \multicolumn{1}{c}{[3, 200]} & \multicolumn{1}{c}{[20, 200]} & \multicolumn{1}{c}{[50, 200]} & \multicolumn{1}{c}{[3, 200]} & \multicolumn{1}{c}{[20, 200]} & \multicolumn{1}{c}{[50, 200]} \\
    \midrule
    NGMN-GCN (Our)                    & 97.73$\rpm$0.11   & 98.29$\rpm$0.21   & 96.81$\rpm$0.96       & 96.56$\rpm$0.12   & 97.60$\rpm$0.29   & 92.89$\rpm$1.31 \\
    \midrule
    \tabincell{c}{NGMN-GraphSAGE}     & 97.31$\rpm$0.56       & 98.21$\rpm$0.13       & 97.88$\rpm$0.15           & 96.13$\rpm$0.30       & 97.30$\rpm$0.72       & 93.66$\rpm$3.87 \\
    \tabincell{c}{NGMN-GIN}           & 97.97$\rpm$0.08       & 98.06$\rpm$0.22	    & 94.66$\rpm$4.01           & 96.98$\rpm$0.20       & 97.42$\rpm$0.48	    & 92.29$\rpm$2.23 \\
    \tabincell{c}{NGMN-GGNN}          &\bf 98.42$\rpm$0.41    & \bf 99.77$\rpm$0.07   & \bf 97.93$\rpm$1.18       & \bf 99.35$\rpm$0.06   & \bf 98.51$\rpm$1.04   & \bf94.17$\rpm$7.74    \\
    \bottomrule
    \end{tabular}
\end{table*}

\subsubsection{Impact of Different Numbers of Perspectives}
We further investigate the impact of different numbers of perspectives adopted by the node-graph matching layer in NGMN.
Following the same settings of previous experiments, we only change the number of perspectives (\ie, $\widetilde{d} = 50/75/100/125/150$) of NGMN.
From Table~\ref{tab:ablation:classification_different_perspectives},
it is clearly seen that the AUC score of NGMN does not increase as the number of perspectives grows for the graph-graph classification task.
Similar results of the graph-graph regression task can be found in Table~\ref{tab:ablation:appendix:regression_different_perspectives} in the Appendix.
We thus conclude that the performance of NGMN is not sensitive to the number of perspectives $\widetilde{d}$ (from 50 to 150) and we make $\widetilde{d}=100$ by default.

\subsubsection{Impact of Different GNN Variants}
We investigate the impact of different GNN variants including GraphSAGE~\cite{GraphSage:hamilton2017inductive}, Graph Isomorphism Network~(GIN)~\cite{xu2018powerful}, and Gated Graph Neural Network~(GGNN)~\cite{li2016gated} adopted by the node embedding layer of NGMN for both the graph-graph classification and graph-graph regression tasks.
Table~\ref{tab:ablation:classification_othergnn} presents the results of the graph-graph classification task (see Table~\ref{tab:ablation:appendix:regression_othergnn} in the Appendix for the results of the graph-graph regression task).
In general, the performance of different GNNs is quite similar for all datasets of both tasks, which indicates that NGMN is not sensitive to the choice of GNN variants in the node embedding layer.
An interesting observation is that NGMN-GGNN performs even better than our default NGMN-GCN on both \textbf{FFmpeg} and \textbf{OpenSSL} datasets.
This shows that our models can be further improved by adopting more advanced GNN models or choosing the most appropriate GNNs according to different real-world application tasks.

\subsubsection{Impact of Different Numbers of GCN Layers}
\begin{table*}[htbp]
    \centering
    \caption{The graph-graph classification results of NGMN models with different numbers of GCN layers in terms of AUC scores (\%).}
    \label{tab:ablation:classification_different_gcn}
      \renewcommand\tabcolsep{16pt}
      \begin{tabular}{lcccccc}
      \toprule
      \multicolumn{1}{c}{\multirow{3}{*}{Model}} & \multicolumn{3}{c}{\bf FFmpeg} & \multicolumn{3}{c}{\bf OpenSSL} \\
      \cmidrule(lr){2-4} \cmidrule(lr){5-7}
      \multicolumn{1}{c}{} & \multicolumn{1}{c}{[3, 200]} & \multicolumn{1}{c}{[20, 200]} & \multicolumn{1}{c}{[50, 200]} & \multicolumn{1}{c}{ [3, 200]} & \multicolumn{1}{c}{ [20, 200]} & \multicolumn{1}{c}{[50, 200]} \\
      \midrule
      NGMN-(1 layer)  & 97.84$\rpm$0.08    & 71.05$\rpm$2.98   & 75.05$\rpm$17.20  & 97.51$\rpm$0.24   & 88.87$\rpm$4.79   &   77.72$\rpm$7.00 \\
      NGMN-(2 layers)  & \bf98.03$\rpm$0.15    & 84.72$\rpm$12.60  & 90.58$\rpm$10.12  & \bf97.65$\rpm$0.10   & 95.78$\rpm$3.46   &     86.39$\rpm$8.16 \\
      NGMN-(3 layers)  & 97.73$\rpm$0.11    & \bf98.29$\rpm$0.21   & 96.81$\rpm$0.96   & 96.56$\rpm$0.12   & 97.60$\rpm$0.29   &    92.89$\rpm$1.31  \\
      NGMN-(4 layers)  & 97.96$\rpm$0.22    & 98.06$\rpm$0.13   & \bf97.94$\rpm$0.15   & 96.79$\rpm$0.21   & \bf98.21$\rpm$0.31   & \bf93.40$\rpm$1.78 \\
      \bottomrule
    \end{tabular}
 \end{table*}

We also examine how the number of GNN (\ie, GCN) layers would affect the performance of our models for both the graph-graph classification and graph-graph regression tasks.
Follow the same default experimental settings, we only change the number of GCN layers in the node embeddings layer of NGMN.
Specifically, we change the number of layers from 1, 2, 3, to 4, and summarize the experimental results in Table~\ref{tab:ablation:classification_different_gcn} for the graph-graph classification task as well as Table~\ref{tab:ablation:appendix:regression_different_gcn} in the Appendix for the graph-graph regression task.

It can be observed from Table~\ref{tab:ablation:classification_different_gcn} that the NGMN model with more GCN layers (\ie, 3-layer and 4-layer) provides better and comparatively stable performance for all sub-datasets for both \textbf{FFmpeg} and \textbf{OpenSSL}, while NGMN with fewer GCN layers (\ie, 1-layer or 2-layer) show inferior performance on some sub-datasets.
For instance, NGMN with 1-layer performs extremely poorly on the [20, 200] and [50, 200] sub-datasets of both datasets; NGMN with 2-layer runs poorly on the [20, 200] and [50, 200] sub-datasets of \textbf{FFmpeg} as well as the [50, 200] sub-dataset of \textbf{OpenSSL}.

These observations indicate that the number of GCN layers that are required in our models depends on the different datasets or different tasks.
Thus, to avoid over-tuning this hyper-parameter (\ie, number of GCN layers) on different datasets and tasks as well as take the resource consumption into considerations, we make the three-layer GCN as the default in the node embedding layer for our models.
\section{Related Work}\label{sec:related_work}
\subsection{Conventional Graph Matching}
As introduced in Section~\ref{sec:intro}, the general graph matching can be categorized into \textit{exact} and \textit{error-tolerant} (\ie, inexact) graph matching techniques.
Specifically, exact graph matching techniques aim to find a strict one-to-one correspondence between two (in large parts) identical graphs being matched, while error-tolerant graph matching techniques allow matching between completely non-identical graphs~\cite{riesen2015structural}.
In real-world applications, the constraint of exact graph matching is too rigid such as the presence of noises or distortion in graphs, neglect of node features, and so on. 
Therefore, an amount of work has been proposed to solve the error-tolerant graph matching problem, which is usually quantified by specific similarity metrics,
such as GED~\cite{gao2010survey, riesen2015structural}, maximum common subgraph~\cite{bunke1997relation}, or even more coarse binary similarity, according to different real-world applications.
Particularly for the computation of GED, it is a well-studied NP-hard problem and suffers from exponential computational complexity and huge memory requirements for exact solutions in practice~\cite{mcgregor1982backtrack,zeng2009comparing,blumenthal2020exact}.

\subsection{Graph Similarity Computation}
Considering the great significance and challenge of computing the graph similarity between pairs of graphs, a popular line of research of graph matching techniques focuses on developing approximation methods for better accuracy and efficiency, including traditional heuristic methods~\cite{gao2010survey,riesen2015structural,wu2019scalable,yoshida2019learning} and recent data-driven graph matching networks~\cite{bai2019simgnn,bai2020learning,li2019graph}, as detailed in Section~\ref{subsec:baseline}.

\textbf{Compared with our work}.
Our work belongs to the data-driven graph matching networks, but theoretically differs from prior work in three main aspects:
1) Unlike prior work only consider either graph-level or node-level interactions, we propose a new type of cross-level node-graph interactions in NGMN to more effectively exploit different-level granularity features between two graphs;
2) Our full model \system combines the advantages of both SGNN and NGMN to capture both global level graph-graph interaction features and cross-level node-graph interaction features between two graphs;
3) Existing work evaluates the graph similarity tasks by only considering either graph-graph classification or regression tasks, while our evaluation is conducted on both two types of tasks and shows superior performance than existing work.

\section{Conclusion and Future Work}\label{sec:conclusion_future_work}
In this paper, we presented a novel multilevel graph matching network (\system) for computing the graph similarity between any pair of graph-structured objects in an end-to-end fashion.
In particular, we further proposed a new node-graph matching network for effectively learning cross-level interactions between two graphs beyond low-level node-node and global-level graph-graph interactions.
Our extensive experimental results correlated the superior performance and robustness compared with state-of-the-art baselines on both the graph-graph classification and graph-graph regression tasks.
One interesting future direction is to either explore other interactions (\eg, subgraph-graph interactions, subgraph-subgraph interactions, etc) for the graph similarity learning or adapt our model \system for solving different real-world applications such as malware detection, text entailment, and question answering with knowledge graphs.
Another highly challenging direction is to study adversarial attacks~\cite{zugner2018adversarial,ling2019deepsec,zhang2020adversarial} against the graph matching/similarity models and further build more robust models.


%




\ifCLASSOPTIONcaptionsoff
  \newpage
\fi

\bibliographystyle{IEEEtran}
\bibliography{reference}

%

\begin{IEEEbiography}[{\includegraphics[width=1in,height=1.25in,clip,keepaspectratio]{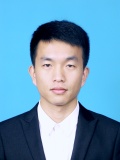}}]{Xiang Ling}
received the B.E. degree from the China University of Petroleum (East China) in 2015.
He is currently pursuing the Ph.D. degree with the College of Computer Science and Technology, Zhejiang University.
His research mainly focus on the data-driven security, AI security and graph neural network.
\end{IEEEbiography}

\begin{IEEEbiography}[{\includegraphics[width=1.0in,height=1.25in,clip,keepaspectratio]{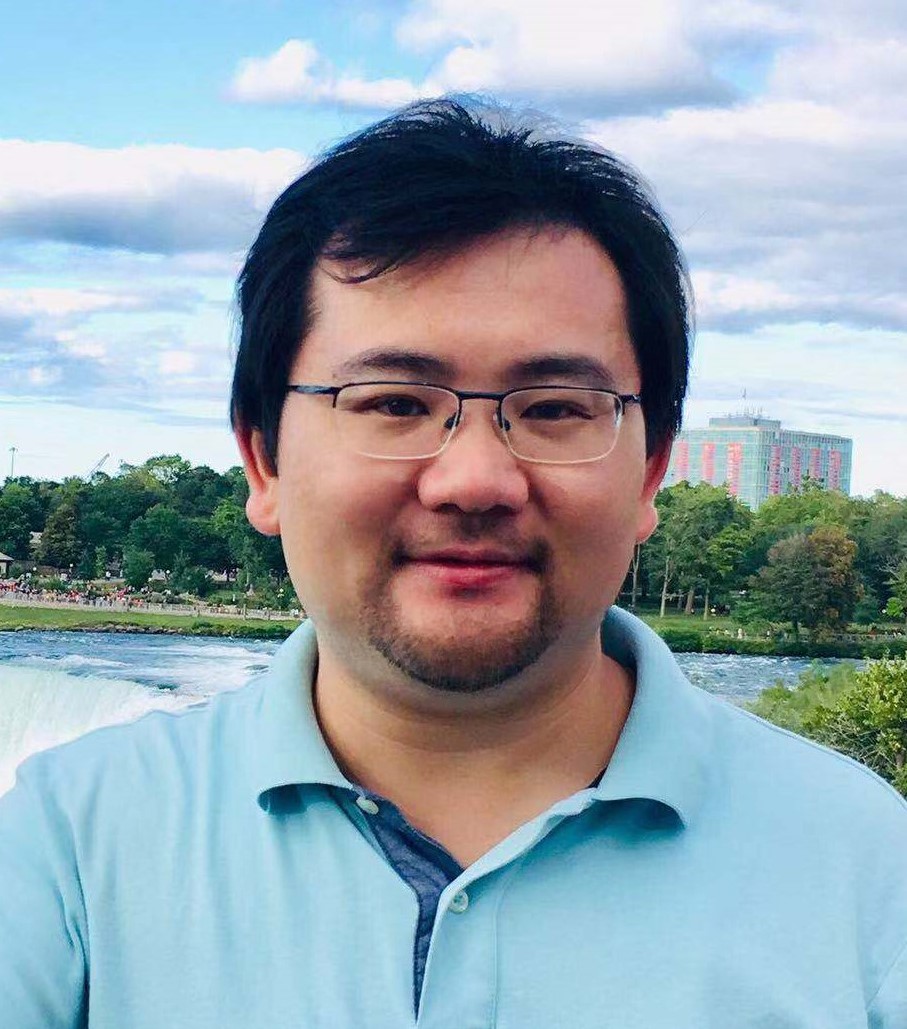}}]{Lingfei Wu} is a Principal Scientist at JD.COM Silicon Valley Research Center, leading a team of 30+ machine learning/natural language processing scientists and software engineers to build intelligent ecommerce personalization systems. He earned his Ph.D. degree in computer science from the College of William and Mary in 2016. Previously, he was a research staff member at IBM Thomas J. Watson Research Center and led a 10+ research scientist team for developing novel Graph Neural Networks methods and systems, which leads to the \#1 AI Challenge Project in IBM Research and multiple IBM Awards including three-time Outstanding Technical Achievement Award. He has published more than 90 top-ranked conference and journal papers, and is a co-inventor of more than 40 filed US patents. Because of the high commercial value of his patents, he has received several invention achievement awards and has been appointed as IBM Master Inventors, class of 2020. He was the recipients of the Best Paper Award and Best Student Paper Award of several conferences such as IEEE ICC’19, AAAI workshop on DLGMA’20 and KDD workshop on DLG'19. His research has been featured in numerous media outlets, including NatureNews, YahooNews, Venturebeat, and TechTalks. He has co-organized 10+ conferences (KDD, AAAI, IEEE BigData) and is the founding co-chair for Workshops of Deep Learning on Graphs (with AAAI’21, AAAI’20, KDD’20, KDD’19, and IEEE BigData’19). He has currently served as Associate Editor for  IEEE Transactions on Neural Networks and Learning Systems, ACM Transactions on Knowledge Discovery from Data and International Journal of Intelligent Systems, and regularly served as a SPC/PC member of the following major AI/ML/NLP conferences including KDD, IJCAI, AAAI, NIPS, ICML, ICLR, and ACL.
\end{IEEEbiography}

\begin{IEEEbiography}[{\includegraphics[width=1in,height=1.25in,clip]{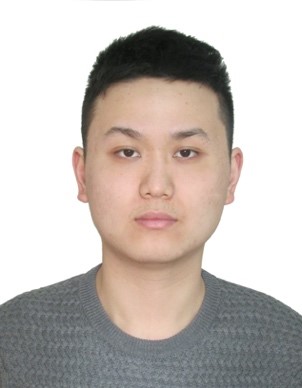}}]{Saizhuo~Wang}
received his B.E. degree in Computer Science and Technology from Zhejiang University at Hangzhou, China in 2020. He is now pursuing his Ph.D. degree in Computer Science and Engineering at Hong Kong University of Science and Technology.
His research mainly focuses on natural language processing and deep learning.
\end{IEEEbiography}

\begin{IEEEbiography}[{\includegraphics[width=1in,height=1.25in,clip,keepaspectratio]{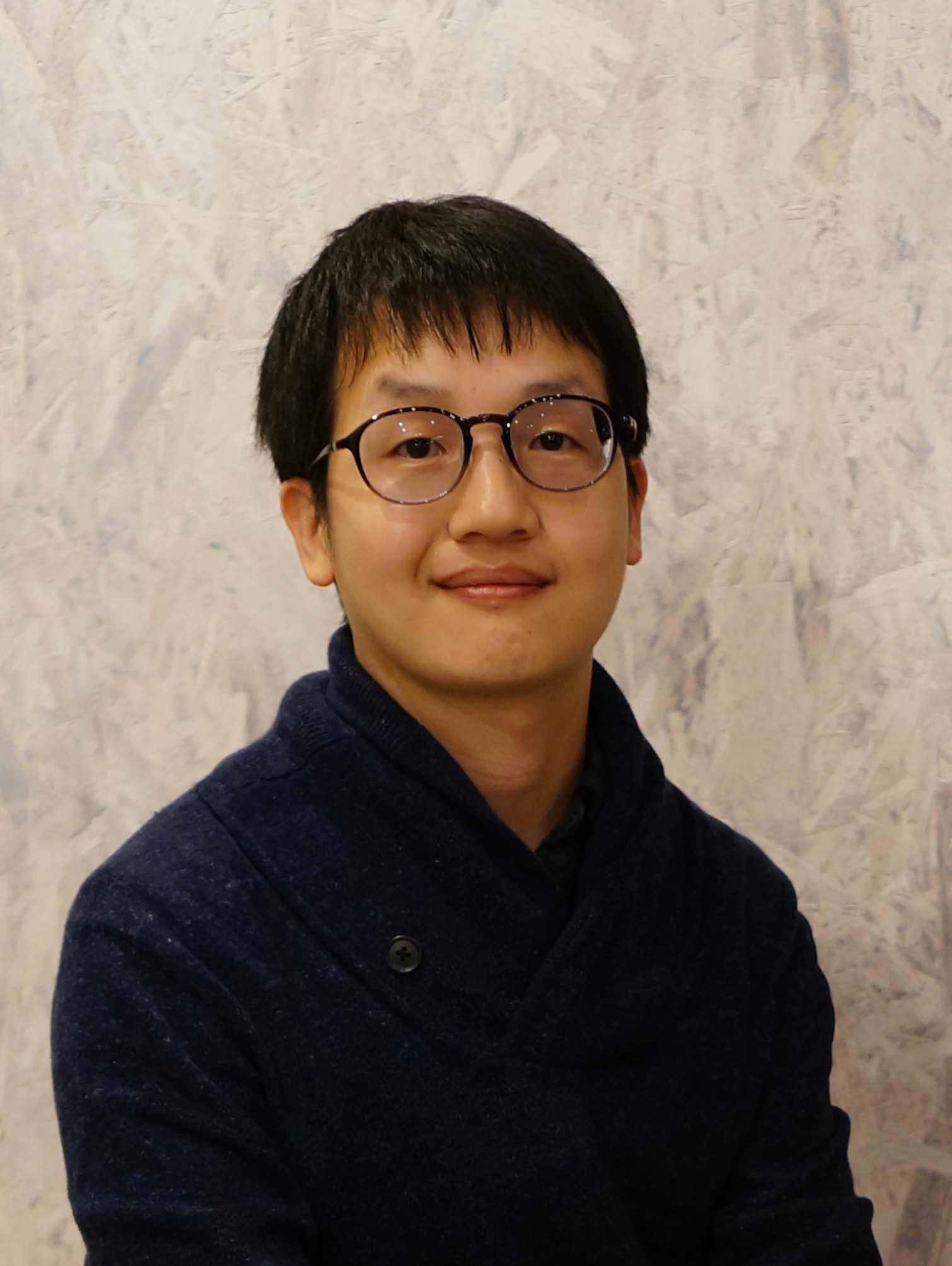}}]{Tengfei~Ma}
is a research staff member of IBM Research AI.
Prior to joining IBM T. J. Watson Research Center, he obtained his Ph.D. from the University of Tokyo and worked in IBM Research Tokyo for one year.
Before that he got his master’s degree from Peking University and his bachelor's degree from Tsinghua University.
His research interests have spanned a number of different topics in machine learning and natural language processing.
Particularly, his recent research is focused on graph neural networks and their applications.
\end{IEEEbiography}

\begin{IEEEbiography}[{\includegraphics[width=1in,height=1.25in,clip,keepaspectratio]{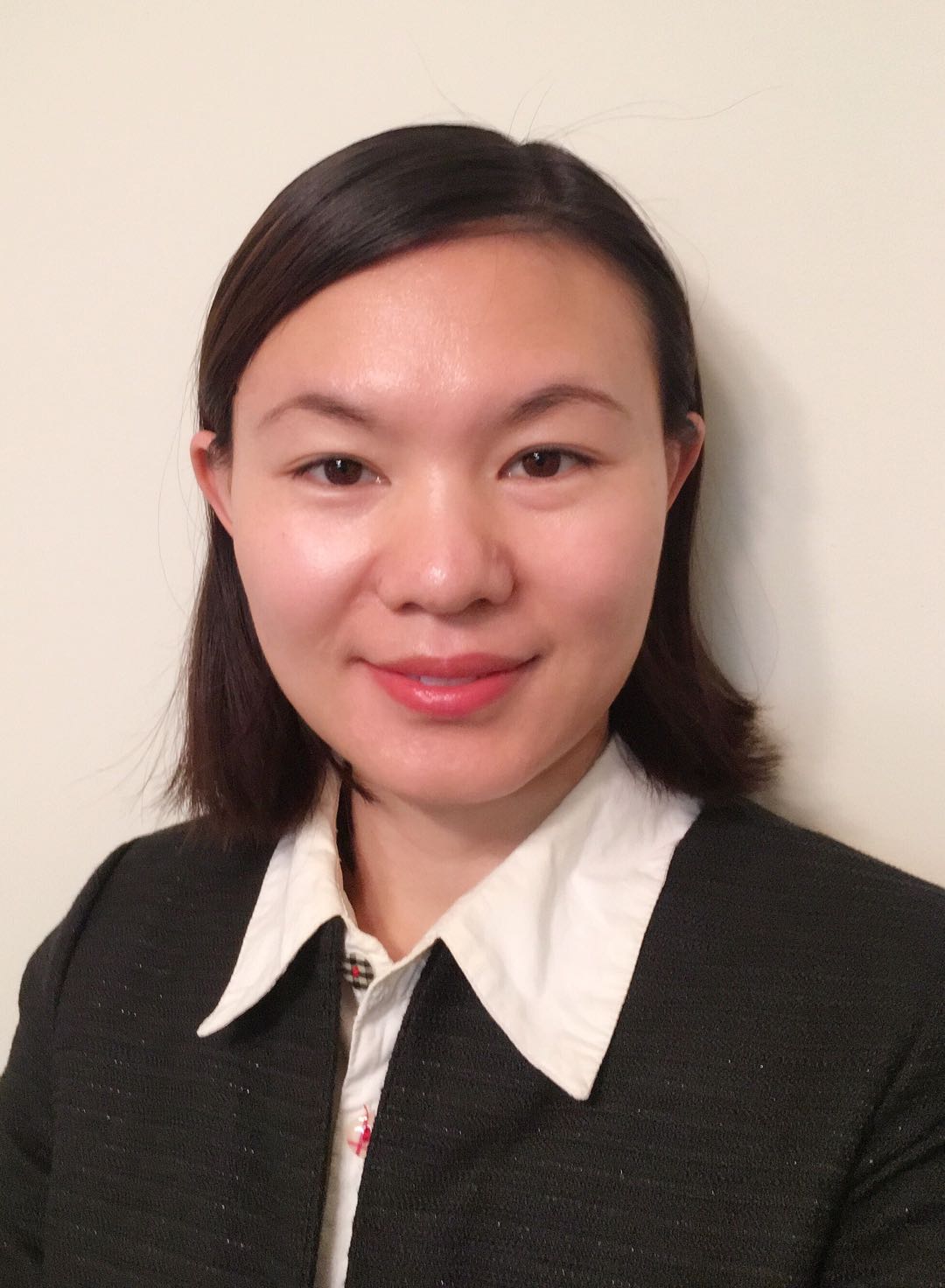}}]{Fangli Xu} earned her master degree in computer science from the College of William and Mary in 2018. She is a research engineer at Squirrel AI Learning. Her research interests mainly focus on machine learning, deep learning, and natural language processing, with a particular focus on AI + Education. 
\end{IEEEbiography}

\begin{IEEEbiography}[{\includegraphics[width=1in,height=1.25in,clip]{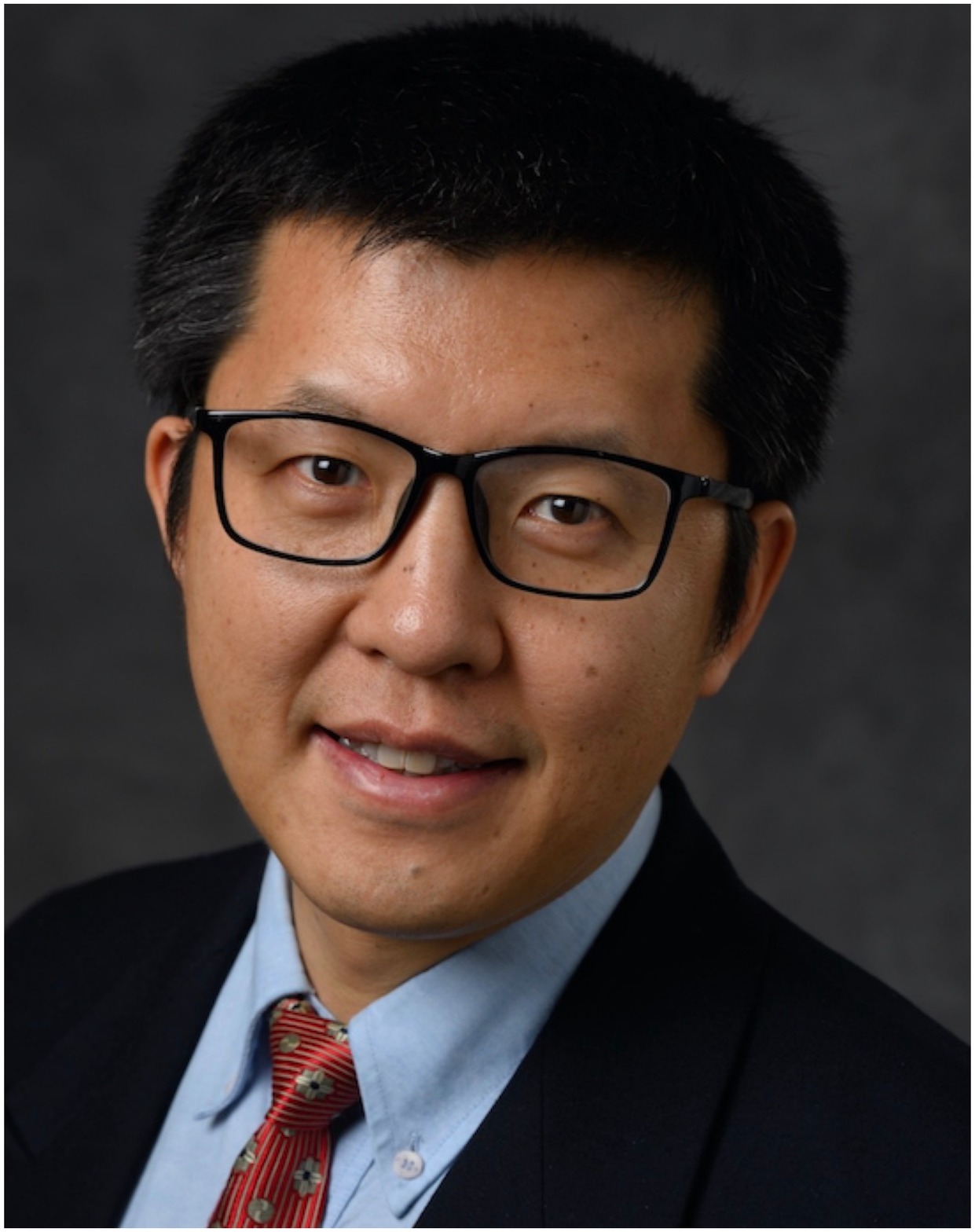}}]{Alex~X.~Liu}
(Fellow, IEEE)
received his Ph.D. degree in Computer Science from The University of Texas at Austin in 2006, and is currently the Chief Scientist of the Ant Group, China. Before that, he was a Professor of the Department of Computer Science and Engineering at Michigan State University. 
He received the IEEE \& IFIP William C. Carter Award in 2004, a National Science Foundation CAREER award in 2009, the Michigan State University Withrow Distinguished Scholar (Junior) Award in 2011, and the Michigan State University Withrow Distinguished Scholar (Senior) Award in 2019.
He has served as an Editor for IEEE/ACM Transactions on Networking, and he is currently an Associate Editor for IEEE Transactions on Dependable and Secure Computing, IEEE Transactions on Mobile Computing, and an Area Editor for Computer Communications.
He has served as the TPC Co-Chair for ICNP 2014 and IFIP Networking 2019.
He received Best Paper Awards from SECON-2018, ICNP-2012, SRDS-2012, and LISA-2010.
His research interests focus on networking, security, and privacy.
He is an IEEE Fellow and an ACM Distinguished Scientist.
\end{IEEEbiography}

\begin{IEEEbiography}[{\includegraphics[width=1in,height=1.25in,clip]{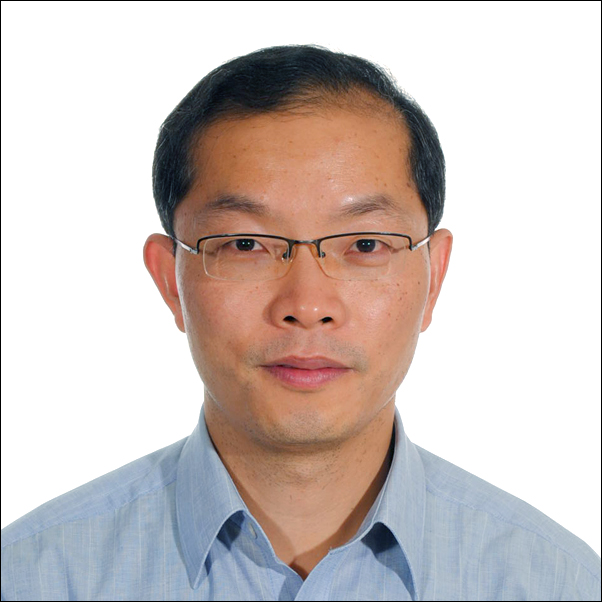}}]{Chunming~Wu}
received the Ph.D. degree in computer science from Zhejiang University, Hangzhou, China, in 1995.
He is currently a Professor with the College of Computer Science and Technology, Zhejiang University.
His research interests include software defined networks, proactive network defense, network virtualization, and intelligent networks.
\end{IEEEbiography}

\begin{IEEEbiography}[{\includegraphics[width=1in,height=1.25in,clip,keepaspectratio]{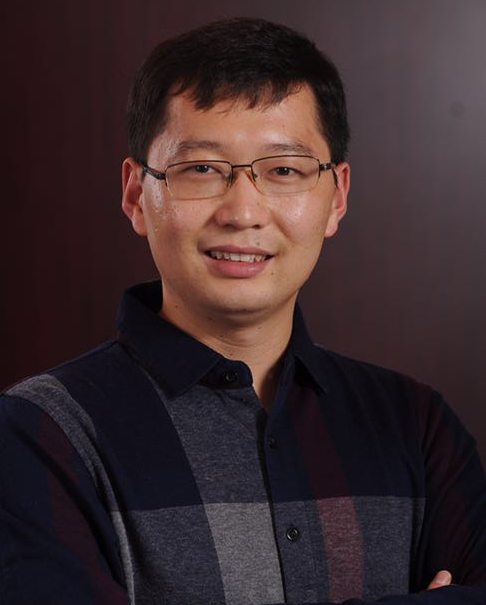}}]{Shouling~Ji}
is a ZJU 100-Young Professor in the College of Computer Science and Technology at Zhejiang University and a Research Faculty in the School of Electrical and Computer Engineering at Georgia Institute of Technology (Georgia Tech).
He received a Ph.D. degree in Electrical and Computer Engineering from Georgia Institute of Technology , a Ph.D. degree in Computer Science from Georgia State University, and B.S. (with Honors) and M.S. degrees both in Computer Science from Heilongjiang University.
His current research interests include Data-driven Security and Privacy, AI Security and Big Data Analytics.
He is a member of ACM, IEEE, and CCF and was the Membership Chair of the IEEE Student Branch at Georgia State University (2012-2013).
He was a Research Intern at the IBM T. J. Watson Research Center.
Shouling is the recipient of the 2012 Chinese Government Award for Outstanding Self-Financed Students Abroad.
\end{IEEEbiography}







\appendix
\subsection{More Details about Datasets}\label{appendix:more_datasets}
For \textbf{FFmpeg}, we prepare the CFGs as the benchmark dataset to detect binary function similarity.
First, we compile \emph{FFmpeg 4.1.4} using 2 different compilers (\ie, \emph{gcc 5.4.0} and \emph{clang 3.8.0}) and 4 different compiler optimization levels (\emph{O0-O3}), which produce a total of 8 different types of compiled binary files.
Second, these 8 generated binaries are disassembled with IDA Pro\footnote{IDA Pro, \url{https://www.hex-rays.com/products/ida/index.shtml}.}, which can produce CFGs for all disassembled functions.
Finally, for each basic block in CFGs, we extract 6 block-level numeric features as the initial node representation based on IDAPython (a python-based plugin in IDA Pro).
\textbf{OpenSSL} is built from OpenSSL (v1.0.1f and v1.0.1u) using \emph{gcc 5.4} in three different architectures (x86, MIPS, and ARM), and four different optimization levels (\emph{O0-O3}).
The \textbf{OpenSSL} dataset that we evaluate is previously released by~\cite{xu2017neural} and publicly available\footnote{\url{https://github.com/xiaojunxu/dnn-binary-code-similarity}.} with prepared 6 block-level numeric features.

Overall, for both \textbf{FFmpeg} and \textbf{OpenSSL} datasets, each node in the CFGs are initialized with 6 block-level numeric features as follows:
\emph{\# of string constants}, \emph{\# of numeric constants}, \emph{\# of total instructions}, \emph{\# of transfer instructions}, \emph{\# of call instructions}, and \emph{\# of arithmetic instructions}.

\subsection{Detailed Experimental Settings for Baseline Models}\label{appendix:baseline_settings}
In principle, we follow the same experimental settings as the baseline methods in their original papers and adjust a few settings to fit specific tasks.
For instance, \textbf{SimGNN} is originally used for the graph-graph regression task, we modify the final layer of the model architecture so that it can be used to evaluate the  graph-graph classification task fairly.
To be specific, detailed experimental settings of all three baselines for both the graph-graph classification and graph-graph regression tasks are given as follows.

\textbf{SimGNN}:
SimGNN firstly adopts a three-layer GCN to encode each node of a pair of graphs into a vector.
Then, SimGNN employs a two-stage strategy to model the similarity between the two graphs:
i) it uses the neural tensor network (NTN) to interact two graph-level embedding vectors that are aggregated by a node attention mechanism;
ii) it uses the histogram features extracted from the pairwise node-node similarity scores.
Finally, the features learned from the two-stage strategy are concatenated to feed into multiple fully connected layers to obtain a final prediction.

For the graph-graph regression task, the output dimensions for the three-layer GCN are 64, 32, and 16, respectively.
The number of K in NTN and the number of histogram bins are both set to 16.
Four fully connected layers are employed to reduce the dimension of concatenated results from 32 to 16, 16 to 8, 8 to 4, 4 to 1.
As for training, the loss function of mean square error is used to train the model with the Adam optimizer.
The learning rate is set to 0.001 and the batch size is set to 128.
We set the number of iterations to 10,000 and select the best model based on the lowest validation loss.

To fairly compare our models with SimGNN in evaluating the graph-graph classification task, we adjust the settings of SimGNN as follows.
We follow the same architecture of SimGNN in the graph-graph regression task except that the output dimension of the last connected layer is set to 2.
We apply a Softmax operation over the output of SimGNN to get the predicted binary label for the graph-graph classification task.
As for training, we use the cross-entropy loss function to train the model and set the number of epochs to 100.
Other training hyper-parameters are kept the same as the graph-graph regression task.

\textbf{GMN}:
The spirit of GMN is improving the node embeddings of one graph by incorporating the implicit neighbors of another graph through a soft attention mechanism.
GMN follows a similar model architecture of the neural message-passing network with three components:
an encoder layer that maps the node and edge to initial vector features of node and edge,
a propagation layer further update the node embeddings through proposed strategies, and an aggregator that computes a graph-level embedding vector for each graph.

For the graph-graph classification task, we use a 1-layer MLP as the node/edge encoder and set the number of rounds of propagation to 5.
The dimension of the node embedding is set to 32, and the dimension of graph-level representation vectors is set to 128.
The Hamming distance is employed to compute the distance of two graph-level representation vectors.
Based on the Hamming distance, we train the model with the margin-based pairwise loss function for 100 epochs in which validation is carried out per epoch.
The Adam optimizer is used with a learning rate of 0.001 and a batch size of 10.

To enable fair comparisons with GMN for the graph-graph regression task, we adjust the GMN by concatenating the graph-level representation of two graphs and feeding it into a four-layer fully connected layers like SimGNN so that the final output dimension is reduced to 1.
As for training, we use the mean square loss function with batch size 128.
Other settings remain the same as the graph-graph classification task.

\textbf{GraphSim}:
The main idea of GraphSim is to convert the graph similarity computation problems into pattern recognition problems.
GraphSim first employs GCN to generate node embeddings of a pair of input graphs, then turns the two sets of node embeddings into a similarity matrix consisting of the pairwise node-node interactions, feeds these matrices into convolutional neural networks (CNN), and finally concatenates the results of CNN to multiple fully connected layers to obtain a final predicted graph-graph similarity score.

For the graph-graph regression task, three layers of GCN are employed with each output dimension being set to 128, 64, and 32, respectively.
The following architecture of CNNs is used:
$Conv(6, 1, 1, 16)$, $Max(2)$,
$Conv(6, 1, 16, 32)$, $Max(2)$,
$Conv(5, 1, 32, 64)$, $Max(2)$,
$Conv(5, 1, 64, 128)$, $Max(3)$,
$Conv(5, 1, 128, 128)$, $Max(3)$.
Numbers in $Conv()$ represent the window size, kernel stride, input channels, and output channels of the CNN layer, respectively.
The number in $Max()$ denotes the pooling size of the max pooling operation.
Eight fully connected layers are used to reduce the dimension of the concatenated results from CNNs, from 384 to 256, 256 to 128, 128 to 64, 64 to 32, 32 to 16, 16 to 8, 8 to 4, 4 to 1.
As for training, the loss function of mean square error is used to train the model with Adam optimizer.
The learning rate is set to 0.001 and the batch size is set to 128.
Similar to SimGNN, we set the number of iterations to 10,000 and select the best model based on the lowest validation loss.

To make a fair comparison of our models with GraphSim in our evaluation, we also adjust GraphSim to solve the graph-graph classification task.
We follow the same architecture of GraphSim in the graph-graph regression task except that seven connected layers are used instead of eight.
The output dimension of the final connected layer is set to 2 and we apply a Softmax operation over it to get the predicted binary label for the graph-graph classification task.
As for training, we use the cross-entropy loss function to train our models and set the number of epochs to 100.
Other training hyper-parameters are kept the same as the graph-graph regression task.

\begin{table*}[htbp]
  \centering
  \caption{The graph-graph regression results of NGMN models with different numbers of perspectives in terms of $mse$, $\rho$, $\tau$, $p@10$ \& $p@20$.}
  \label{tab:ablation:appendix:regression_different_perspectives}
    \renewcommand\tabcolsep{14.5pt}
    \begin{tabular}{clccccc}
    \toprule
    Datasets & \multicolumn{1}{c}{Model} & $mse$ ($10^{-3}$) & $\rho$ & $\tau$ & $p@10$ & $p@20$ \\
    \midrule
    \multicolumn{1}{c}{\multirow{5}[0]{*}{\textbf{\tabincell{c}{AIDS700}}}}
    & \tabincell{c}{NGMN-($\widetilde{d}=~50$)} &   \bf1.133$\rpm$0.044    &   \bf0.909$\rpm$0.001    &   \bf0.756$\rpm$0.002    &   \bf0.487$\rpm$0.006    &   \bf0.563$\rpm$0.007  \\
    & \tabincell{c}{NGMN-($\widetilde{d}=~75$)} &   1.181$\rpm$0.053    &   0.905$\rpm$0.005    &   0.750$\rpm$0.007    &   0.468$\rpm$0.026    &   0.547$\rpm$0.025  \\
    & \tabincell{c}{NGMN-($\widetilde{d}=100$)} &   1.191$\rpm$0.048    &   0.904$\rpm$0.003    &   0.749$\rpm$0.005    &   0.465$\rpm$0.011    &   0.538$\rpm$0.007 \\
    & \tabincell{c}{NGMN-($\widetilde{d}=125$)} &   1.235$\rpm$0.062    &   0.900$\rpm$0.007    &   0.743$\rpm$0.010    &   0.456$\rpm$0.021    &   0.531$\rpm$0.014 \\
    & \tabincell{c}{NGMN-($\widetilde{d}=150$)} &   1.301$\rpm$0.059    &   0.893$\rpm$0.005    &   0.734$\rpm$0.007    &   0.435$\rpm$0.021    &   0.511$\rpm$0.022 \\
    \midrule
    \multicolumn{1}{c}{\multirow{5}[0]{*}{\textbf{\tabincell{c}{LINUX1000}}}}
    & \tabincell{c}{NGMN-($\widetilde{d}=~50$)} &   \bf1.260$\rpm$0.070    &   \bf0.954$\rpm$0.004    &   \bf0.829$\rpm$0.007    &   0.825$\rpm$0.021    &   0.823$\rpm$0.025    \\
    & \tabincell{c}{NGMN-($\widetilde{d}=~75$)} &   1.330$\rpm$0.108    &   0.952$\rpm$0.003    &   0.826$\rpm$0.006    &   \bf0.833$\rpm$0.029    &   \bf0.843$\rpm$0.035    \\
    & \tabincell{c}{NGMN-($\widetilde{d}=100$)} &   1.561$\rpm$0.020    &   0.945$\rpm$0.002    &   0.814$\rpm$0.003    &   0.743$\rpm$0.085    &   0.741$\rpm$0.086    \\
    & \tabincell{c}{NGMN-($\widetilde{d}=125$)} &   1.406$\rpm$0.184    &   0.950$\rpm$0.006    &   0.823$\rpm$0.015    &   0.799$\rpm$0.111    &   0.803$\rpm$0.068    \\
    & \tabincell{c}{NGMN-($\widetilde{d}=150$)} &   1.508$\rpm$0.083    &   0.946$\rpm$0.003    &   0.815$\rpm$0.005    &   0.756$\rpm$0.033    &   0.758$\rpm$0.027    \\
    \bottomrule
    \end{tabular}
\end{table*}
\begin{table*}[htbp]
  \centering
  \caption{The graph-graph regression results of NGMN models with different GNNs in terms of $mse$, $\rho$, $\tau$, $p@10$ \& $p@20$.}
  \label{tab:ablation:appendix:regression_othergnn}
  \renewcommand\tabcolsep{13.5pt}
    \begin{tabular}{clccccc}
    \toprule
    Datasets & \multicolumn{1}{c}{Model} & $mse$ ($10^{-3}$) & $\rho$ & $\tau$ & $p@10$ & $p@20$ \\
    \midrule
    \multicolumn{1}{c}{\multirow{4}[0]{*}{\textbf{\tabincell{c}{AIDS700}}}}
    & NGMN-GCN (Our)                & \bf1.191$\rpm$0.048   &\bf0.904$\rpm$0.003    &\bf0.749$\rpm$0.005      &\bf0.465$\rpm$0.011    &\bf0.538$\rpm$0.007      \\
    \cmidrule{2-7}
    & \tabincell{c}{NGMN-(GraphSAGE)}     & 1.275$\rpm$0.054      &	0.901$\rpm$0.006    &	0.745$\rpm$0.008      &	0.448$\rpm$0.016	  & 0.533$\rpm$0.014      \\
    & \tabincell{c}{NGMN-(GIN)}           & 1.367$\rpm$0.085	    & 0.889$\rpm$0.008	  & 0.729$\rpm$0.010	    & 0.400$\rpm$0.022	  & 0.492$\rpm$0.021      \\
    & \tabincell{c}{NGMN-(GGNN)}          & 1.870$\rpm$0.082	    & 0.871$\rpm$0.004	  & 0.706$\rpm$0.005	    & 0.388$\rpm$0.015	  & 0.457$\rpm$0.017       \\
    \midrule
    \multicolumn{1}{c}{\multirow{4}[0]{*}{\textbf{\tabincell{c}{LINUX1000}}}}
    & NGMN-GCN (Our)                 & 1.561$\rpm$0.020      & 0.945$\rpm$0.002       & 0.814$\rpm$0.003      & 0.743$\rpm$0.085      & 0.741$\rpm$0.086 \\
    \cmidrule{2-7}
    & \tabincell{c}{NGMN-GraphSAGE}  & 2.784$\rpm$0.705	   & 0.915$\rpm$0.019	    & 0.767$\rpm$0.028	    & 0.682$\rpm$0.183	    & 0.693$\rpm$0.167    \\
    & \tabincell{c}{NGMN-GIN}        & \bf1.126$\rpm$0.164   & \bf0.963$\rpm$0.006	& \bf0.858$\rpm$0.015	&\bf0.792$\rpm$0.068    &\bf0.821$\rpm$0.035    \\
    & \tabincell{c}{NGMN-GGNN}       & 2.068$\rpm$0.991	   & 0.938$\rpm$0.028	    & 0.815$\rpm$0.055	    & 0.628$\rpm$0.189	    & 0.654$\rpm$0.176    \\
    \bottomrule
    \end{tabular}
\end{table*}
\begin{table*}[htbp]
    \centering
    \caption{The graph-graph regression results of NGMN models with different numbers of GCN layers in terms of $mse$, $\rho$, $\tau$, $p@10$ \& $p@20$.}
    \label{tab:ablation:appendix:regression_different_gcn}
    \renewcommand\tabcolsep{14.5pt}
      \begin{tabular}{clccccc}
      \toprule
      Datasets & \multicolumn{1}{c}{Model} & $mse$ ($10^{-3}$) & $\rho$ & $\tau$ & $p@10$ & $p@20$ \\
      \midrule
      \multicolumn{1}{c}{\multirow{4}{*}{\textbf{\tabincell{c}{AIDS700}}}}
      & NGMN-(1 layer)  &  1.297$\rpm$0.025    &   0.895$\rpm$0.001    &   0.737$\rpm$0.002    &   0.414$\rpm$0.011    &   0.498$\rpm$0.006 \\
      & NGMN-(2 layers)  &  \bf1.127$\rpm$0.015    &   \bf0.908$\rpm$0.001    &   \bf0.755$\rpm$0.002    &   \bf0.479$\rpm$0.009    &   \bf0.555$\rpm$0.006 \\
      & NGMN-(3 layers)  &  1.191$\rpm$0.048    &   0.904$\rpm$0.003    &   0.749$\rpm$0.005    &   0.465$\rpm$0.011    &   0.538$\rpm$0.007 \\
      & NGMN-(4 layers)  &  1.345$\rpm$0.098    &   0.887$\rpm$0.009    &   0.727$\rpm$0.012    &   0.401$\rpm$0.034    &   0.491$\rpm$0.029 \\
      \midrule
      \multicolumn{1}{c}{\multirow{4}{*}{\textbf{\tabincell{c}{LINUX1000}}}}
      & NGMN-(1 layers)  &  \bf1.449$\rpm$0.234    &   0.943$\rpm$0.013    &   0.817$\rpm$0.018    &   0.750$\rpm$0.070    &   \bf0.786$\rpm$0.065 \\
      & NGMN-(2 layers)  &  1.525$\rpm$0.119    &   \bf0.948$\rpm$0.003    &   \bf0.818$\rpm$0.005    &   0.706$\rpm$0.076    &   0.736$\rpm$0.039 \\
      & NGMN-(3 layers)  &  1.561$\rpm$0.020    &   0.945$\rpm$0.002    &   0.814$\rpm$0.003    &   0.743$\rpm$0.085    &   0.741$\rpm$0.086 \\
      & NGMN-(4 layers)  &  1.677$\rpm$0.248    &   0.943$\rpm$0.008    &   0.810$\rpm$0.013    &   \bf0.758$\rpm$0.063    &   0.765$\rpm$0.071 \\
      \bottomrule
      \end{tabular}
\end{table*}

\end{document}